\documentclass[10pt,twocolumn,twoside]{IEEEtran}
\usepackage{cite}
\newcommand\citep{\cite}

\ifCLASSINFOpdf
   \usepackage[pdftex]{graphicx}
\else
   \usepackage[dvips]{graphicx}
\fi
\usepackage{epsfig}
\usepackage[usenames, dvipsnames]{color}
\usepackage{subfig}
\usepackage{amsmath}
\usepackage{enumerate}
\usepackage{amsfonts}
\usepackage{comment}
\usepackage{algorithm}
\usepackage{algorithmic}
\usepackage{array}
\usepackage{hyperref}
\usepackage{CJKutf8}

\usepackage{amssymb,bm}
\usepackage{textcomp}
\usepackage{multirow}
\usepackage{tabularx}
\usepackage{mathtools}
\usepackage{array}
\usepackage{hyperref}
\usepackage{hhline}
\newcolumntype{P}[1]{>{\centering\arraybackslash}p{#1}}
\usepackage{diagbox}
\usepackage[singlelinecheck=false]{caption}
\usepackage{placeins}
\usepackage{pifont}
\usepackage{lipsum}

\newcolumntype{L}[1]{>{\raggedright\arraybackslash}m{#1}}
\newcolumntype{N}{@{}m{0pt}@{}}

\begin{document}
\begin{CJK*}{UTF8}{bsmi}
%\title{Scalable Sentiment for Sequence-to-sequence Chatbot Response with Performance Analysis} 
%\title{Investigation of Different Approaches for \\ Sentiment Controllable Chatbot}
\title{Investigation of Sentiment Controllable Chatbot}
\author{\IEEEauthorblockN{Hung-yi Lee, Cheng-Hao Ho, Chien-Fu Lin, Chiung-Chih Chang, \\  Chih-Wei Lee, Yau-Shian Wang, Tsung-Yuan Hsu, Kuan-Yu Chen} \\
\thanks{This work was financially supported by the Ministry of Science and Technology of Taiwan.}
\vspace{1.5ex}
\IEEEauthorblockA{College of Electrical Engineering and Computer Science, National Taiwan University}
} 

\date{College of Electrical Engineering and Computer Science, National Taiwan University}

\maketitle

\begin{abstract}
Conventional seq2seq chatbot models attempt only to find sentences with the highest probabilities conditioned on the input sequences, without considering the sentiment of the output sentences. In this paper, we investigate four models to scale or adjust the sentiment of the chatbot response: a persona-based model, reinforcement learning, a plug and play model, and CycleGAN, all based on the  seq2seq model. We also develop machine-evaluated metrics to estimate whether the responses are reasonable given the input. These metrics, together with human evaluation, are used to analyze the performance of the four models in terms of different aspects;  reinforcement learning and CycleGAN are shown to be very attractive.
\end{abstract}


\IEEEpeerreviewmaketitle

\section{Introduction}
\label{sec:intro}

In contrast to goal-oriented dialogue systems~\cite{lee2009example,wen2016network}, chatbot chats with human users on any subject domain of daily life~\cite{serban2016building,shang2015neural}.
The conventional chatbot is based on the seq2seq model~\cite{vinyals2015neural}, generating meaningful responses given user input. 
It is usually emotionless, which is a major limitation of modern chatbots as emotion plays a critical role in human social interaction, especially in chatting~\cite{keltner1998emotion}. 
Hence we seek to train the chatbot to generate responses with scalable sentiment by setting the chat mode.
For example, for the input ``How was your day today?'', the chatbot may respond, ``It is wonderful today'' or ``It is terrible today'' depending on the sentiment set, in addition to simply generating a reasonable response.
This mode can either be set by the developer or the user, or determined dynamically based on the dialogue context.
The techniques mentioned here may be extended to conversational style adjustment, so the machine may imitate the conversational style of someone the user is familiar with, to make the chatbot more friendly or more personal~\cite{polzin2000emotion,hasegawa2013predicting}.

Substantial effort has been focused on the conversational fluency and content quality of generated responses, for example, by enriching the content diversity~\cite{vijayakumar2016diverse,li2015diversity,li2016deep}, considering additional information~\cite{li2016persona}, and addressing unknown words~\cite{gu2016incorporating,eric2017copy}. 
Responses have also been generated with controllable factors. 
The sentiment of a given sentence can be modified using non-parallel data~\cite{shen2017style}. 
A chatbot can change the style of responses by optimizing a given sentiment-related function~\cite{mueller2017sequence}. 
However, little work has been reported on scaling the sentiment of a chatbot; it remains difficult to evaluate a chatbot with adjustable sentiment properly~\cite{shawar2007different,hung2009towards}. 

In this paper, we investigate four approaches to scale the sentiment of chatbot responses and use a set of evaluation metrics and human evaluation with which we analyze the approaches.
This journal paper is an extension of a conference paper\cite{SentimentLee2018}, but with additional results on two more corpora.

\section{Related Work}
%The approaches used in this paper are related to controllable sentence generation and text style transfer. 
The approaches presented in Sections~\ref{subsec:persona},~\ref{subsec:reinforce},~\ref{subsec:plugandplay} and~\ref{subsec:cyclegan} are related to Sections~\ref{subsuubsec:relate1},~\ref{subsuubsec:relate2},~\ref{subsuubsec:relate3} and~\ref{subsuubsec:relate4},  respectively. 

\subsection{Controllable Sentence Generation}
Sentence generators based on deep learning show promising text generation capabilities, but cannot easily control the generated text.
Hence, a series of research aims at controlling the generated sentences, for example, the writing styles or the topics of the generated sentences. 

\subsubsection{Controlled by Input Factors} \label{subsuubsec:relate1}
Sentence generation models can take some factors as input to influence the generation of its outputs.
Conditioned recurrent neural networks (CRNN) are used to control the linguistic style of generated text~\cite{ficler-goldberg-2017-controlling}.
Affect-LM customizes the degree of emotional content in generated sentences through an additional design parameter~\cite{AffectLM}.
The conditional transformer language model (CTRL)~\cite{CTRL} trains to condition on control codes that govern style, content, and task-specific behavior. 
%Transformer is equipped to achieve better style transfer and content preservation in~\cite{dai2019style}. %Lee: 還是放到 text style transfer 比較好

This category of approach has been used in dialogue generation.
The persona model~\cite{li2016persona} encodes personas in distributed embeddings that capture individual characteristics such as background information and speaking style, and the embeddings influence the output of the decoder. 
Instead of encoding personas,  in Section~\ref{subsec:persona}, the persona-based model is used to control sentiment. 
A conversational model is proposed to generate informative responses with controlled sentence function, for example, interrogative, imperative, declarative, etc~\cite{sentence_function}. %Does this approach belong to this type?
This paper is closely related to the Emotional Chatting Machine (ECM)~\cite{zhou2017emotional}.
ECM is a neural conversational model that can generate corresponding responses based on given emotional categories.
The basic idea of ECM  is similar to the persona-based model, but with more sophisticated network architectures, including internal and external memories.
However, it needs the dialogues involving emotional responses to train the model, which is not always available. 

\subsubsection{Controlled by External Function} \label{subsuubsec:relate2}
In this approach, the sentence generation model is explicitly taught to generate sentences with certain aspects~\cite{ECIR_affect,Uber_PaP,RLarXiv19,SentiGAN}. 
%\cite{Uber_PaP} %where can we put this paper
%~\cite{ECIR_affect}. %Add extra loss, like RL approach % used in dialogue
A hand-crafted or machine-learned function guide the sentence generator to output sentences considered having the desired aspect (like sentiment or topic) based on the function. 
This approach has been used to make the dialogue generation model generate emotional responses, but hand-crafted functions are used in the previous work~\cite{ECIR_affect}.  
It has been shown that the attribute models learned from data can successfully guide the sentence generator~\cite{Uber_PaP}. 
Because here we focus on dialogue generation, in Section~\ref{subsec:reinforce}, besides considering the attribute, or sentiment, of the responses, we further guide the model to generate the responses coherent with the input sentences by coherence models. 

\subsection{Text Style Transfer} 
The text style transfer model transfers the input sentence from one style into another. 
The text style transfer approaches that do not utilize parallel data are used in Sections~\ref{subsec:plugandplay} and~\ref{subsec:cyclegan}.
Below are two main categories of approaches to achieve text style transfer without parallel data.
All the related work mentioned below only focuses on text style transfer, not dialogue generation as in Section~\ref{sec:approach}.

\subsubsection{Manipulating Latent Space}\label{subsuubsec:relate3}
The latent representations of auto-encoders can be manipulated to induce a change in the output space to achieve text style transfer~\cite{pmlr-v80-zhao18b,mueller2017sequence}.
The approach used in Section~\ref{subsec:plugandplay} belongs to this category.
One way to manipulate latent space to achieve text style transfer is feature disentangle.
By separating the content from the style in the latent space, we can modify the style without changing the content~\cite{shen2017style,Fu2017StyleTI,disentangleNAACP19,disentangleAAAI2020}.
%The adversarial network is used to align sentences with different styles without parallel data~\cite{shen2017style}, and it can be used to disentangle content representations and style representations~\cite{Fu2017StyleTI}.

\subsubsection{Direct Modification}\label{subsuubsec:relate4}
Instead of manipulating the latent space, this category of approach directly finds a model that can transform the text from one style to another. 
The approach used in Section~\ref{subsec:cyclegan} belongs to this category.
A simple approach is to delete phrases associated with the sentence’s style and retrieve new phrases to replace them~\cite{ruleNAACL18}. 
The idea similar to CycleGAN~\cite{zhu2017unpaired} or StarGAN~\cite{StarGAN}, which has widely used in image style transfer, has also been used.
This category of approaches uses a discriminator to control the style of the generated content~\cite{NIPS2018_7959,cycleGANACL19}, and reconstruction loss to maintain the content~\cite{rewriting_ICLR19}. 
\section{Sentiment Controllable Chatbot}
\label{sec:approach}
In Section~\ref{subsec:seq2seq} we briefly review the conventional seq2seq chatbot.
The four approaches used here are presented in Sections~\ref{subsec:persona} to~\ref{subsec:cyclegan}.
All use the seq2seq chatbot as the basic model.
The persona-based approach (Section~\ref{subsec:persona}) and reinforcement learning (Section~\ref{subsec:reinforce}) modify the training algorithm of the conventional seq2seq chatbot.
Plug and play (Section~\ref{subsec:plugandplay}) and CycleGAN (Section~\ref{subsec:cyclegan}) modify the response of an off-the-shelf seq2seq chatbot.
Below we assume that the chatbot response is to be positive conditioned on the input, although it is simple to generalize the approaches to scalable sentiment.

\subsection{Seq2seq Model} \label{subsec:seq2seq}
\begin{figure}[h]
        \centering
        \includegraphics[width=\linewidth]{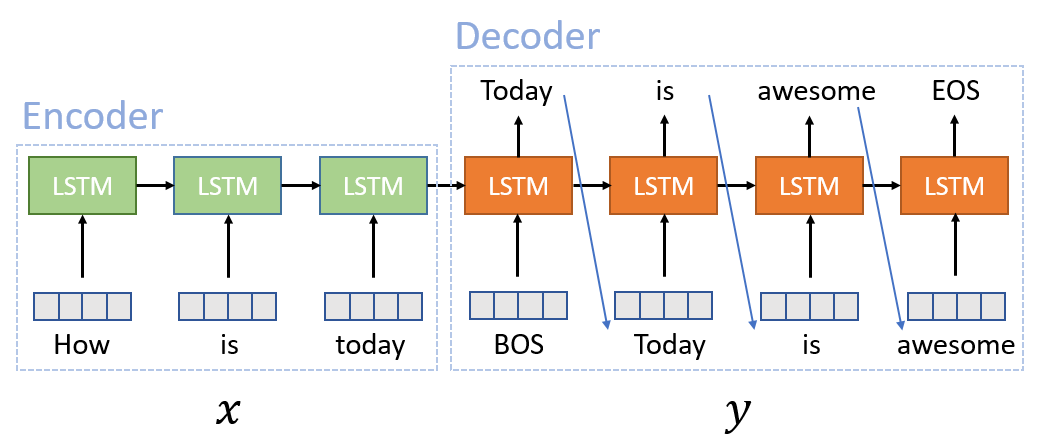}
        \caption{{\it Seq2seq model}}
        \label{fig:seq2seq}
\end{figure}

%\vspace{-1mm}
Here we use the attention-based seq2seq model~\cite{luong2015effective} shown in Figure~\ref{fig:seq2seq} to train a simple chatbot using a corpus of dialogue pairs.
In all discussions here, $x$ is the input sentence to the seq2seq chatbot, and $y$ is the output of the seq2seq model.
$\hat{y}$ is the reference response in the training corpus.
In the training phase, we input the sentence $x$ (a sequence of one-hot vectors) to the encoder, and the seq2seq model learns to maximize the probability of generating the sentence $\hat{y}$ given $x$. 

%\vspace{-4mm}
%\subsection{The Four Proposed Approaches} \label{subsec:five}
\subsection{Persona-Based Model} \label{subsec:persona}
%\vspace{-1mm}

\begin{figure}[h]
        \centering
        \includegraphics[width=\linewidth]{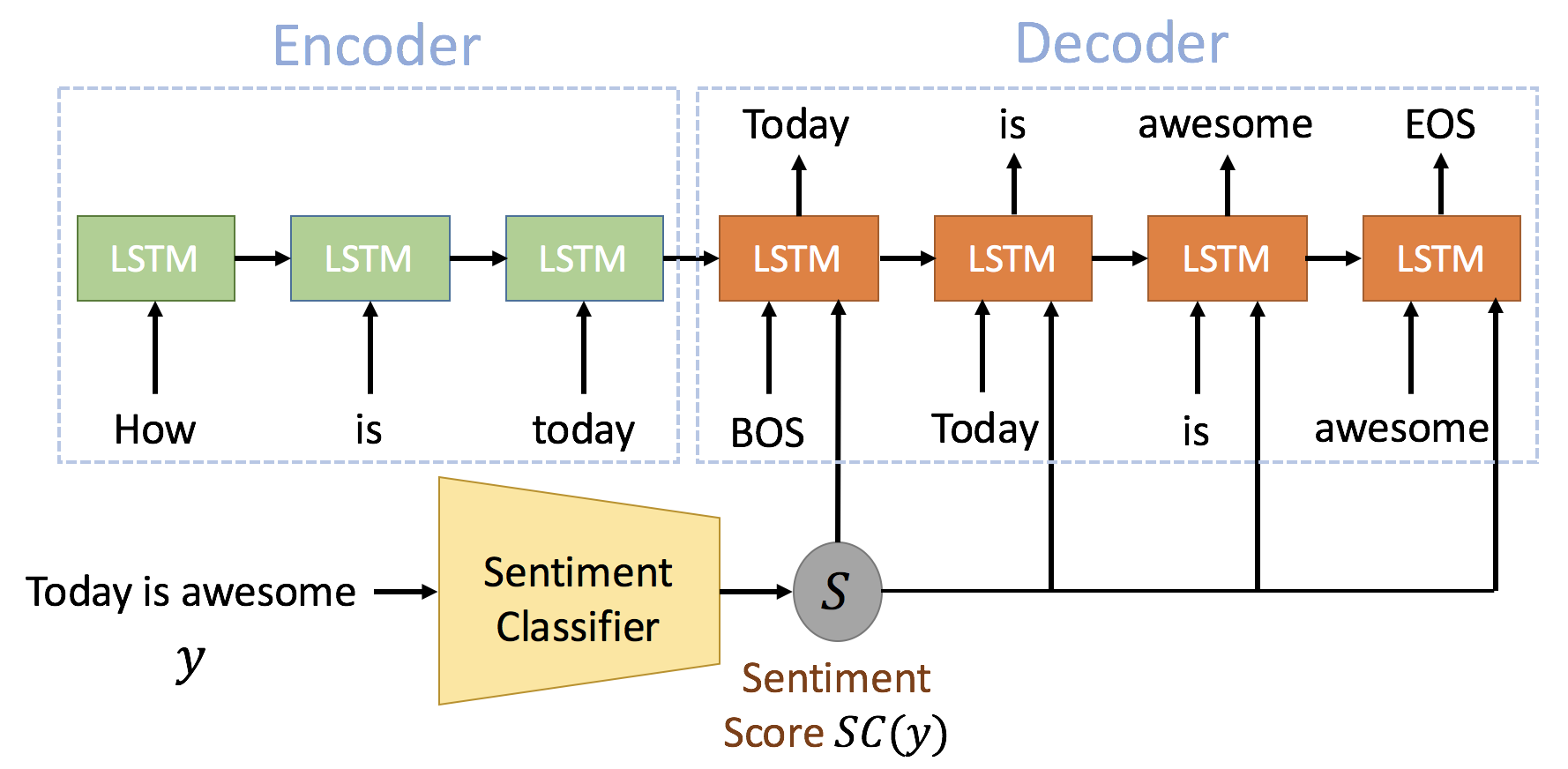}
        \caption{{\it Persona-based Seq2seq model.}}
        \label{fig:persona}
\end{figure}

The persona-based model was originally proposed to generate sentences that mimic the responses of specific speakers~\cite{li2016persona}.
It is very similar to the seq2seq model, except that extra information is added to the input of the decoder at each time step. 
In the original work~\cite{li2016persona}, this extra information is the trained speaker embedding.
Here we replace the speaker embedding with a sentiment score (a scalar between $0$ and $1$) from a sentiment classifier, as shown in Figure~\ref{fig:persona}.
This sentiment classifier~\cite{liu2012sentiment} is trained on a corpus of sentences with labeled sentiments to determine whether a sentence is positive or not.
The input of the classifier is a sentence $z$, and the output is a score $SC(z)$ between $0$ and $1$ indicating how positive the input is.
The input of the decoder at every time step is then the concatenation of the word embedding and a sentiment score.
During training the sentiment score of the reference sentence $SC(\hat{y})$ is used, and the decoder learns to generate the reference sentence.
For testing given the same input, we scale the sentiment of the output by entering the desired sentiment score.

\subsection{Reinforcement Learning} \label{subsec:reinforce}
%\vspace{-1mm}
Here we use exactly the seq2seq chatbot shown in Figure~\ref{fig:seq2seq}; the only modification is a set of reward functions designed to scale the response sentiment using reinforcement learning.
The components of the reward functions are developed as follows.

\begin{enumerate}
    \item  \textit{Semantic Coherence 1}:
In addition to being a good sentence, the response $y$ should be semantically relevant to the input $x$.
Hence we pre-train a different seq2seq model on a large dialogue corpus to estimate this semantic coherence with a probability $P_{coh}(y|x)$.
The first reward is therefore
\begin{equation}\label{equ:lm}
R_1 = \frac{1}{N_y}\cdot log P_{coh}(y|x),
\end{equation}
where $x$ and $y$ denote the input and response of the baseline seq2seq chatbot (not the pre-trained seq2seq model), and $N_y$ is the length of $y$ for normalization.

\item \textit{Semantic Coherence 2}: 
The semantic coherence mentioned above can be estimated in a completely different way.
We use the same dialogue corpus to train a RNN discriminator, in which two RNN encoders are used to represent the input $x$ and its corresponding response $y$ as two embeddings; these two embeddings are concatenated and followed by a fully connected layer to produce a score $D_{\mathit{RNN}}(x,y)$ between $0$ and $1$ which indicates whether $x$ and $y$ are good dialogue pairs.
This score is therefore the second reward:
%\vspace{-1mm}
\begin{equation}\label{equ:ch}
R_2 = D_{\mathit{RNN}}(x,y).
\end{equation}

\item \textit{Sentiment Score}: 
The third reward is based on the sentiment classifier mentioned in Section~\ref{subsec:persona}:
%\vspace{-1mm}
\begin{equation}\label{equ:sc}
R_3 = SC(y),
\end{equation}
where $y$ is the seq2seq chatbot response.

\end{enumerate}

The total reward is then the linear interpolation of the three rewards mentioned above:
\begin{equation}
R = \alpha \cdot R_1 + \beta \cdot R_2 + (1-\alpha-\beta) \cdot R_3,
\end{equation}
where $\alpha$ and $\beta$ are hyper-parameters ranging from $0$ to $1$ and $\alpha$ $+$ $\beta$ $<$ 1.
We employ the reinforcement learning algorithm with policy gradient~\cite{sutton2000policy}.

\subsection{Plug and Play Model} \label{subsec:plugandplay} 
%\vspace{-1mm}

\begin{figure}[h]
        \centering
        \includegraphics[width=\linewidth]{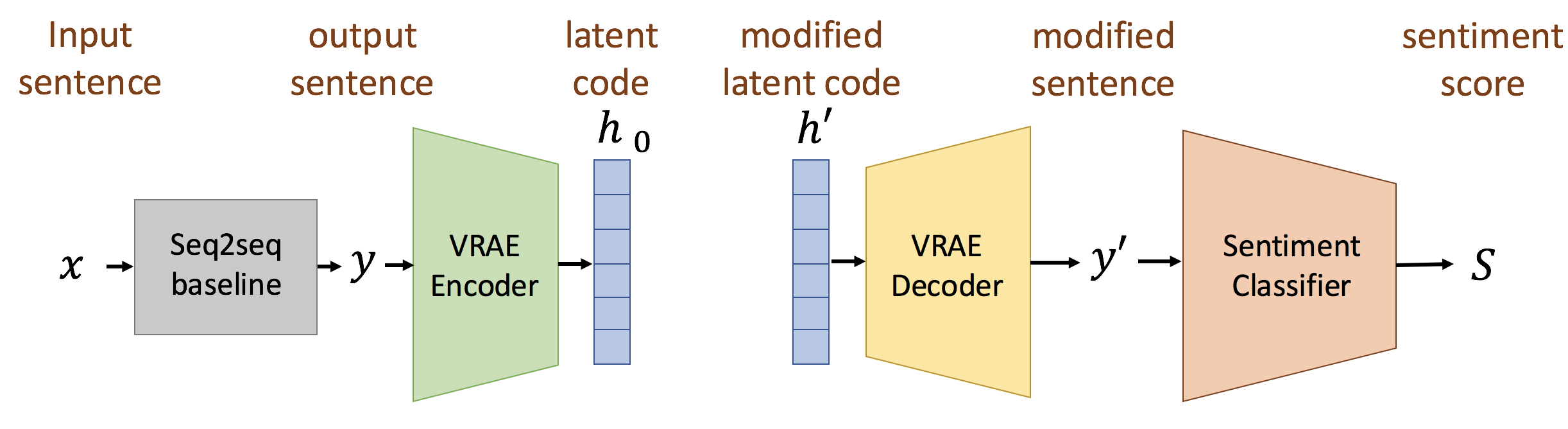}
        \caption{{\it Plug and play model. VRAE denotes variational recurrent auto-encoder.}}
        \label{fig:plugandplay}
\end{figure}
%\vspace{-3mm}
As shown in Figure \ref{fig:plugandplay}, to generate dialogue responses here, we borrow the concept of plug and play as used in generating images~\cite{nguyen2016plug}.
Here we pre-train a variational recurrent auto-encoder (VRAE)~\cite{fabius2014variational} in addition to using the same dialogue corpus. 
The VRAE encoder on the left transforms a sentence $y$ into a fixed-length latent vector $h_0$, while the VRAE decoder on the middle right generates a sentence $y'$ based on a vector $h'$. 
The VRAE encoder and decoder are also jointly learned from the dialogue corpus for the chatbot.

\textit{The following steps take place on-line, when the user enters a sentence.}
Given an input $x$, the seq2seq baseline first generates a response $y$ which is then encoded into a latent code $h_0$ by the VRAE encoder. 
Then the latent code $h_0$ is modified into $h^\prime$, based on the following equation:
\begin{equation}
h' = \mathit{argmax}_{h} [\gamma\cdot SC( Decoder(h) ) - \delta\cdot MSE(h,h_0)],
\label{eq:pap}
\end{equation}
%\hspace{-1mm}
where $SC$ denotes the sentiment classifier and $\gamma$ and $\delta$ are the weights of the loss function term and the regularization term.
The first term on the right-hand side of Eq.~(\ref{eq:pap}) indicates that we seek a code $h$ such that when decoded into a sentence $Decoder(h)$ using the VRAE decoder, the resulting sentiment score $SC(Decoder(h))$ is maximized.
The second term of Eq.~(\ref{eq:pap}) prevents the code $h'$ from drifting too far from $h_0$.
To solve Eq.~(\ref{eq:pap}), we calculate the gradient of the sentiment score with respect to the latent code $h$ and apply gradient ascent to the latent code iteratively, until the sentiment score output reaches a pre-defined value.
Because Eq.~(\ref{eq:pap}) is solved on-line after the user enters an input sentence, this approach is more time consuming.
Since the argmax layer between the decoder and sentiment classifier in $SC(Decoder(h))$ is non-differentiable, we use soft argmax~\cite{kusner2016gans} to approximate argmax so that the gradient back-propagates throughout the whole network, from the sentiment classifier to the decoder.

\subsection{CycleGAN} \label{subsec:cyclegan}
%\vspace{-1mm}
%put figure here...
\begin{figure}[h]
        \centering
        \includegraphics[width=\linewidth]{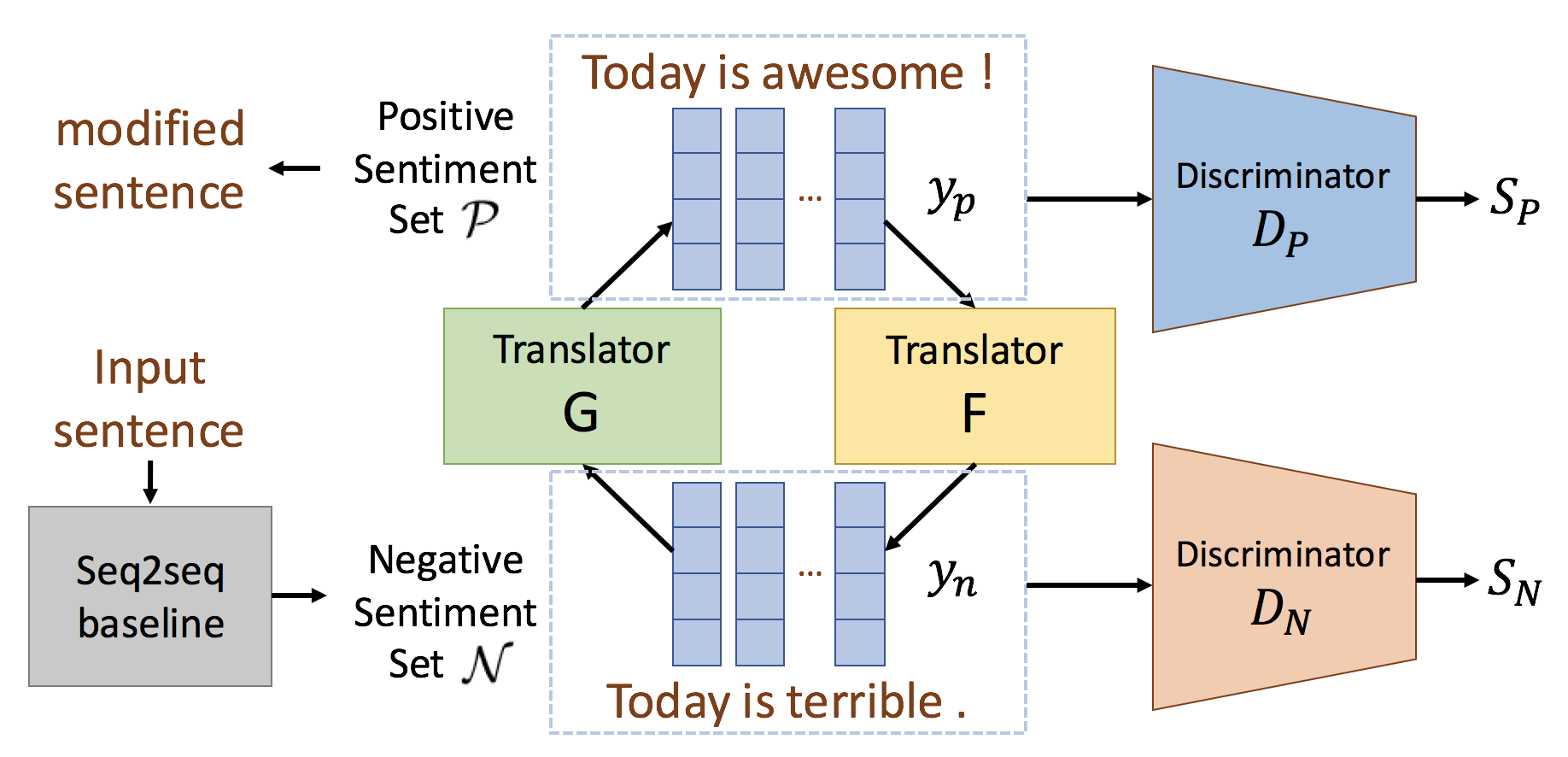}
        \caption{{\it CycleGAN model for sentiment transformation. $F$ and $G$ are two translators respectively from positive to negative and negative to positive, and $D_P$ and $D_N$ are two discriminators respectively for positive and negative sentiment.}}
        \label{fig:cycleGAN}
\end{figure}

Here we adopt the very powerful cycle generative adversarial network (CycleGAN), which proved successful in image style transformation even without paired data~\cite{zhu2017unpaired}. 
As illustrated in Figure~\ref{fig:cycleGAN}, we show a way to use CycleGAN to transform the sentiment of sentences from negative to positive. 
The model is trained on two sets of sentences in a corpus with labeled sentiments: a positive sentiment set $\mathcal{P}$ and a negative sentiment set $\mathcal{N}$. 
The sentences in the two sets are unpaired; that is, for a given sentence in $\mathcal{P}$, it is not known which is the corresponding sentence in $\mathcal{N}$.
We train two seq2seq translators: $G$ to transform a negative sentence $y_n$ to positive and $F$ for positive $y_p$ to negative.
We also train discriminators $D_{\mathcal{P}}$ and $D_{\mathcal{N}}$.
They take a sequence of word embeddings as input and learn to distinguish whether the sequence is from word embeddings of a real sentence, or it was generated by $G$ or $F$.
With the continuous word embeddings as the translator output, the gradient can be back-propagated from the discriminator to the translator. 
Note that $F$ and $G$ transform sequences of word embeddings to sequences of word embeddings.  % AMH: is this correct? yes
We pre-train the word embedding model with Word2Vec~\cite{mikolov2013efficient}; here it is fixed during CycleGAN training. 
To transform the output sequence of word embeddings into a sentence, we simply select those words whose embeddings have the highest cosine-similarity to each given word embedding in the sequence.

The concept of W-GAN~\cite{arjovsky2017wasserstein} is used to train $D_{\mathcal{P}}$ and $D_{\mathcal{N}}$.
The loss function of the discriminator $D_{\mathcal{P}}$ is
%\vspace{-1mm}
\begin{equation}
L(D_{\mathcal{P}}) = D_{\mathcal{P}}(G(y_n)) - D_{\mathcal{P}}(y_p), \label{eq:dp}
\end{equation}
where $y_n$ is a negative sentence sampled from $\mathcal{N}$, and $G(y_n)$ is the output of translator $G$ taking $y_n$ as the input.
$D_{\mathcal{P}}$ learns to minimize Eq.~(\ref{eq:dp}), that is, to give low scores to the translated output (the first term on the right) and high scores to real positive sentences $y_p$ (the second term).
The loss function of the discriminators $D_{\mathcal{N}}$ is parallel to Eq.~(\ref{eq:dp}):
%\vspace{-1mm}
\begin{equation}
L(D_{\mathcal{N}}) = D_{\mathcal{N}}(F(y_p)) - D_{\mathcal{N}}(y_n). \label{eq:dn}
\end{equation}

As in improved W-GAN, gradient penalty is applied here.
The loss functions for training translators $G$ and $F$ are
%\vspace{-1mm}
\begin{equation}
\begin{aligned}
L(F) &= 2[MSE(y_p, G(F(y_p))) + MSE(y_n, F(G(y_n)))] \\
&\qquad - D_{\mathcal{N}}(F(y_p)), \label{eq:generatorG}
\end{aligned}
\end{equation}
%\vspace{-1mm}
\begin{equation}
\begin{aligned}
L(G) &= 2[MSE(y_p, G(F(y_p))) + MSE(y_n, F(G(y_n)))] \\
&\qquad - D_{\mathcal{P}}(G(y_n)). \label{eq:generatorF}
\end{aligned}
\end{equation}

The first terms on the right-hand side of Eqs.~(\ref{eq:generatorG}) and (\ref{eq:generatorF}) are the same.
Given a positive sentence $y_p$, after being transformed into a negative sentence by $F$ and then transformed back to positive by $G$, it should be very close to the original sentence $y_p$;
likewise for the second terms.
The last terms of Eqs.~(\ref{eq:generatorG}) and (\ref{eq:generatorF}) are different:
$F$ learns to generate output $F(y_p)$ that is considered by $D_{\mathcal{N}}$ to be a real negative sentence, whereas $G$ learns to generate output $G(y_n)$ that is considered by $D_{\mathcal{P}}$ to be a real positive sentence.
In this way translators $F$ and $G$ learn to transform the sentences from one sentiment (positive or negative) to the other.
Notice that the discriminators $D_\mathcal{P}$ and $D_\mathcal{N}$ are jointly trained with the translators $F$ and $G$.
During testing, for any chatbot output $y$, we simply use $G$ to transform it into a positive sentence $G(y)$.

\section{Experimental Setup}\label{sec:experiment} 
We trained and tested all our models, including the seq2seq and the four proposed models, on the following three corpora. %both Chinese and English corpus. 
The first two are in Chinese, whereas the third is in English.
%The main experiment is of Chinese corpus, whereas English part is an extra.
Using the training set, we trained five models, including the seq2seq baseline and the four proposed models; we  evaluated these models using the testing set.
All the evaluation metrics obtained are the average over the testing data. %Lee: 為什麼不是 testing??? <-- 李致緯paper原本就是這樣寫的，照他所述這邊的val就是test V
More dataset details are provided in Appendix~\ref{appendix:dataset}.
%Lee: evaluation 到底是用甚麼資料訓練的？應該要註明一下! <-- explained in III V

\begin{enumerate}
\item We used the Chinese Emotional Conversation Generation (CECG) task~\cite{CECG}, originally offered by the NII Testbeds and Community for Information Access Research (NTCIR) Project for the Short Text Conversation Task (STC) competition. %Lee: CECG 應該引用文章 V
CECG contains around 1.7M dialogue pairs; each sentence is labeled by one of the following six kinds of sentiments: like, sad, disgusted, angry, happy, and other. We reclassified five of these six sentiment categories into positive and negative categories: like and happy as positive sentiments, and sad, disgusted, and angry as negative sentiments. 
%Since the CECG corpus is simplified Chinese, we converted it to traditional Chinese in advance. 
Both corpora were split into training and testing sets (the latter included 1k dialogue pairs). 
This corpus was used in both the sentiment classifier and the other models. 
%Lee: train, validation, test 各是多少 <-- 共1.7M,分training & testing, testing佔1k mentioned above V
%Lee: 訓練 sentiment classifier 的資料也跟 chat-bot 用得一樣嗎? V

\item We collected data from the PTT Boy-Girl board containing the titles and all the article replies from page 1 to page 4000, and we used the articles as context and the replies as responses. 
Replies include ``like'' and ``boo'', which roughly correspond to positive and negative sentiment. 
However, as this dataset contains no true dialogue data, and the sentiment is not always precise, we use this dataset only for demonstration and not the main experiment.
As with the previous dataset, this corpus was also used in both the sentiment classifier and other models. 
%Lee: train, validation, test 具體而言各有多少筆? <-- 描述於appendix V
%Lee: sentiment classifier 和前一個 data set 用的是一樣的嗎? <-- no, 描述於VI. EXPERIMENTS ON CHINESE PTT CORPUS (此處只提資料) V

\item For English, the Twitter chatting corpus is available on Marsan-Ma's GitHub repository~\cite{Marsan-Ma} using TensorFlow. 
This corpus, which contains 3.7M dialogue pairs, 
is split into training and testing sets, 
the latter of which includes 28k dialogue pairs.
The sentiment classifier used in this work was trained from the Twitter Sentiment Analysis Corpus~\cite{pak2010twitter}, which consists of 15M data with labeled sentiment ($0$ or $1$).
This corpus was also split into a training and testing set. %Lee:各有多大呢 <-- 描述於appendix v
The trained sentiment classifier achieved an $87\%$ accuracy on the validation set. %Lee: 中文的 sentiment classidier正確率是多少?請列出 solved
\end{enumerate}

\section{Evaluation}
%\vspace{-2mm}

\subsection{Evaluation Metrics}
Evaluation is always difficult in language generation; this is even more so for chatbots.
Here we propose two metrics: semantic coherence 1 and 2 (COH1, COH2) for chatbots, which are scores reflecting the degree to which the output sentence $y$ is a proper response to the input sentence $x$.
These are in fact the semantic coherence 1 and 2 mentioned in Section~\ref{subsec:reinforce} (Reinforcement Learning) designed for the reward function.
However, the seq2seq model and the RNN discriminator used to obtain these two scores are re-trained here and are thus different models.% slightly

The third metric is the sentiment classifier score (SCL) used to measure how positive the output sentence is.
This is the sentiment classifier score $SC(y)$ used in the persona-based model mentioned in Section~\ref{subsec:persona}.
Likewise, the sentiment classifier used here is re-trained and is thus different. % slightly

The fourth metric is the language model score (LM) which measures whether the output sentence $y$ is a good sentence in terms of a language model~\cite{mikolov2010recurrent}.
The language model used here is composed of a two-layer GRU~\cite{cho2014learning} model:
\begin{equation}
LM~Score = \frac{1}{N_y}\cdot log P(y), \label{eq:LM_is_PPL}
\end{equation}
which is the language model probability $P(y)$ for a sentence $y$, normalized with the sentence length $N_y$.
Eq.~(\ref{eq:LM_is_PPL}) is also known as the negative log perplexity (PPL).

Note that SCL and LM, the third and fourth metrics, consider only the output sentence $y$~-- not the input $x$.
COH1 and COH2, the first and second metrics, however, consider the output $y$ given the input $x$.
\textit{In all the following tables, larger evaluation metrics represent better performances.}

\subsection{Metric Models}
The coherence score (1 and 2) in Chinese uses the PTT Gossiping board dialogue corpus collected by Justin Yang~\cite{Justin-Yang}, which contains about 400,000 dialogue pairs. 
Here we split the data into training and testing data sets; the latter contains 1,000 pairs, whereas the other owns the rest. 

For the LM score, in Chinese, we crawled the replies of articles from the top 50 most popular PTT boards, for a total of 
25 million tokens of data.  % AMH: check
In English, the corpus of the One Billion Word Benchmark~\cite{chelba2013one} was used.

%With regard to fairness, we would prefer more heterogeneous data for training the four metric models. However, data resources are limited. Thus except for data used for the coherence score (1 and 2) in Chinese and data for the LM score in both Chinese and English, the left (coherence score for English and sentiment score for both Chinese and English) share the same data with the experiments mentioned in Section~\ref{sec:experiment}. 
The coherence score for English and sentiment score for both Chinese and English share the same data with the experiments mentioned in Section~\ref{sec:experiment}. 
Although the metrics are trained with the same corpus, as mentioned above, the models of the metrics are different because of re-training, which guarantees a certain extent of fairness. 

\section{Individual Models}
\label{sec:individual}
Due to the space limitation, we only show the results of the individual models on Chinese CECG dataset. 
The details of the hyper-parameter setup in the following experiments are shown in Appendix~\ref{appendix:hyper_param}.

\subsection{Sentiment Classifier}
To find a proper sentiment classifier, we evaluated six different ways of segmenting Chinese words. Including word-based and character-based methods, we also evaluated different neural network architectures: a CNN with max pooling (CNN for short), a GRU with the last hidden state output (GRU-last), and a GRU averaging all hidden state outputs for the whole sequence (GRU-avg). The total number of characters used in character-based segmentation was 7,297, and the number of words in word-based segmentation was 50,000. 
We trained these six models with a batch size of 32 and 50,000 epochs.
We evaluated the performance using the accuracy and area under the Receiver Operating Characteristics curve (AUC). 
For the accuracy score, the output scores of the models greater than 0.5 were taken as predictions of positive sentiment; otherwise, they were negative. 
The predictions were then compared to the real answers to calculate the accuracy rate. 
The best results of all architectures shown in Table~\ref{table:cecg_sent_classifier}, GRU-last with word segmentation yielded the best performance on both accuracy and AUC scores. 
GRU-last is then applied in the following experiments as the sentiment classifier.

%\FloatBarrier
\begin{table}[h]
\centering
\footnotesize
\begin{tabular}{|P{0.8cm}|P{0.8cm}|P{0.8cm}|P{0.8cm}|P{0.8cm}|P{0.8cm}|P{0.8cm}|}
\hline
Seg. & \multicolumn{3}{c|}{Word-based} & \multicolumn{3}{c|}{Character-based}\\
\hline
Struc. & CNN & GRU-last & GRU-avg & CNN & GRU-last & GRU-avg\\
\hhline{|=|=|=|=|=|=|=|}
Acc&$\textnormal0.914$&$0.927$&$0.926$&$0.879$&$0.905$&$0.908$\\
\hline 
AUC&$\textnormal0.973$&$0.980$&$0.979$&$0.931$&$0.967$&$0.968$\\
\hline
\end{tabular}
\caption{\it Evaluation of sentiment classifiers under different NN architectures on Chinese CECG dataset.}
\label{table:cecg_sent_classifier}
\end{table}

\begin{comment} %圖就不用了，按照 reviewer 說的
%%\FloatBarrier
\begin{figure}[h]
        \centering
        \includegraphics[width=0.4\textwidth]{acc}
        \caption{{\it Accuracy of sentiment classifiers under different NN architectures on Chinese CECG dataset.}}
        \label{fig:acc}
\end{figure}
%%\FloatBarrier
\begin{figure}[h]
        \centering
        \includegraphics[width=0.4\textwidth]{auc}
        \caption{{\it AUC score of sentiment classifiers under different NN architectures on Chinese CECG dataset.}}
        \label{fig:auc}
\end{figure}
\end{comment}

%Lee:
%1. 請在內文中用文字說明上面兩張圖 V
%2. 圖上的中文請務必拿掉 V
%3. 請說明在以下實驗中最後使用了圖中的那一個模型繼續接下來的實驗 V

\subsection{Persona-Based Model}
With a fixed sentiment score input of 1.0, we first compared the performance between the 64-, 128-, and 256-neuron seq2seq models  % AMH: check
of the persona-based model during the inference step. 
As shown in Table~\ref{table:personal_diff_neurons}, the 256-neuron model scored highest in terms of COH1, COH2, and LM, while the 128-neuron model yielded the best performance on the SCL score.
%Although the 128-neuron model yielded the best performance on the sentiment score, as shown in Table~\ref{table:personal_neurons}, this could be caused by its simple and positive sentences. 
The numbers of neurons did not have a remarkable influence on the results. 
We chose the 256-neuron model in the following experiments.

%Lee: 是不是接下來所有的實驗，seq2seq 都是用 256 neuron 呢? <-- 描述於appendix中 V

%%\FloatBarrier
\begin{table}[h]
\centering
\begin{tabular}{|P{1.5cm}|P{0.5cm}|P{0.9cm}|P{0.9cm}|P{0.9cm}|P{0.9cm}|}
\hline
\multicolumn{2}{|c|}{Metrics} & COH1 & COH2 & SCL & LM\\
\hhline{|=|=|=|=|=|=|}
\multirow{3}{*}{\shortstack[c]{Sentiment\\Score}} & 64 & -9.760 & 0.631 & 0.920 & -2.621 \\
\cline{2-6} 
 & 128 & -9.505 & 0.625 & 0.950 & -2.310\\
\cline{2-6} 
 & 256 & -9.338 & 0.639 & 0.925 & -2.178 \\
\hline 
\end{tabular}
\caption{\it Evaluation of persona-based models of different neuron sizes on Chinese CECG dataset.}
\label{table:personal_diff_neurons}
\end{table}

\begin{comment}
%\FloatBarrier
\begin{table}[h]
\centering
\begin{tabular}{|P{1.8cm}|P{0.5cm}|p{5cm}|}
\hline
\multicolumn{2}{|c|}{\multirow{2}{*}{Input}} & 你每次的希望 都那麼的讓人絕望\\
\multicolumn{2}{|c|}{} & Despair always accompanies your hope\\
\hhline{|=|=|=|}
\multirow{6}{*}{\shortstack[c]{Sentiment\\Score}} & \multirow{2}{*}{64} & 我也是，我也是，我也是很感動的\\
 & & I also feel so, so, so touched\\
\cline{2-3} 
 & \multirow{2}{*}{128} & [哈哈]\\
 & & [haha]\\
\cline{2-3} 
 & \multirow{2}{*}{256} & 我也是很堅強的\\
 & & I will make it through this!\\
\hline
\end{tabular}
\caption{\it Examples generated by persona-based models of different neuron sizes on Chinese CECG dataset.}
\label{table:personal_neurons}
\end{table}
\end{comment}

Further, we applied other sentiment scores as input to generate sentences with different degrees of sentiment. Table~\ref{table:personal_diff_sent} shows the results of 0.0, 0.5, and 1.0 as inputs respectively on the 256-neuron model during inference.
As expected, the lower we set the input, the lower the resultant sentiment scores: this suggests the model is already transforming sentences to different sentiments.

%Lee: 這面這個表格好像放錯了? 這好像是不同 neuron 數目的實驗 <-- 兩個實驗都有(已補上不同neuron數目的實驗) V
%\FloatBarrier
\begin{table}[h]
\centering
\begin{tabular}{|P{1.5cm}|P{0.5cm}|P{0.9cm}|P{0.9cm}|P{0.9cm}|P{0.9cm}|}
\hline
\multicolumn{2}{|c|}{Metrics} & COH1 & COH2 & SCL & LM\\
\hhline{|=|=|=|=|=|=|}
\multirow{3}{*}{\shortstack[c]{Sentiment\\score}} & 0.0 & -7.381 & 0.656 & 0.040 & -1.862\\
\cline{2-6} 
 & 0.5 & -7.777 & 0.659 & 0.443 & -2.682 \\
\cline{2-6} 
 & 1.0 & -9.338 & 0.639 & 0.925 & -2.178\\
\hline 
\end{tabular}
\caption{\it Evaluation of persona-based model under different sentiment score inputs.}
\label{table:personal_diff_sent}
\end{table}

\begin{comment}
The performance of LM score might be affected by the imbalanced data between different sentiment degree counted in table \ref{table:datasize_sent_group}., which data of neutral sentiment score is far less than others. 
%\FloatBarrier
\begin{table}[h]
\centering
\begin{tabular}{|P{1.5cm}|P{1.5cm}|P{1.5cm}|P{1.5cm}|}
\hline
\shortstack[c]{Sentiment\\Score} & [0.0, 0.2] & [0.4, 0.6] & [0.8, 1.0]\\
\hline
\shortstack[c]{Data\\Size} & 564694 & 40279 & 477887\\
\hline 
\end{tabular}
\caption{\it Data size of each sentiment group.}
\label{table:datasize_sent_group}
\end{table}
\end{comment}

Tables~\ref{table:personal_diff_scores} show examples of how different sentiment scores affect the input sentences.

\begin{table*}[htbp]
  \small
  \centering
  \captionsetup{justification=centering}
  \subfloat{%
    \hspace{.1cm}%
    \begin{tabular}{|c|c|p{7.0cm}|p{5.7cm}|}
    \hline
    \multicolumn{2}{|c|}{\multirow{4}{*}{Input}} & [月亮] 今晚出去走走，享受廣州長假最後的安寧。& 終於下班了。今晚非常了得\ [耶]\\
    \multicolumn{2}{|c|}{} & [Moon] Go out for a walk tonight and enjoy the last night of your long vacation in Guangzhou & Finally got off work. Very much tonight [ya]\\
    \hhline{|=|=|=|=|}
    \multirow{6}{*}{\shortstack[c]{Sentiment\\score}}& \multirow{2}{*}{0.0} & 你又要去了？ & 我也是不想上班了？\\
     & & You're going again? & I don't want to go to work either\\
    \cline{2-4} 
     & \multirow{2}{*}{0.5} & 你在哪裏 & 我也是剛下班的？？\\
     & & Where are you? & I just got off work too\\
     \cline{2-4} 
     & \multirow{2}{*}{1.0} & 好開心喔 & 恭喜恭喜\\
     & & Sounds like fun & Congratulations\\
    \hline 
    \end{tabular}
    \hspace{.1cm}%
  }
  \caption{\it Sentences generated by persona-based model under different sentiment scores on Chinese CECG dataset.}
  \label{table:personal_diff_scores}
\end{table*}

\subsection{Reinforcement Learning}
In this experiment, different coefficient reward combinations were adopted to test the performance of the model. Below we report the results after 2500 training epochs.
In the first three sets of experiments, we fixed $R_2$ as 0.0 and adjusted the proportion of $R_1$ and $R_3$ to determine how the four metrics were affected. The results in Table~\ref{table:RL_diff_weight_2500} show that the COH1 score increases as $R_1$ rises from 0.0 to 0.8, and the SCL score falls as $R_3$ decreases from 1.0 to 0.2, which is not surprising as rewards $R_1$, $R_2$, and $R_3$ in Eqs.~(\ref{equ:lm}), (\ref{equ:ch}), and (\ref{equ:sc}) were in parallel with the COH1, COH2, and SCL score respectively. Interestingly, COH2 degrades compared to the pretrained MLE baseline, showing that the RL model is highly goal-oriented~-- to improve COH1 and SCL, it sacrifices COH2 performance. The result also implies that COH1 and COH2 are nearly mutually independent as the increase of $R_1$ seems to little affect COH2. 
The LM score, on the other hand, generally is improved comparing with the typical seq2seq model: this is perhaps attributable to $R_1$, which also takes into account word ordering. 
In the last set of experiments, we increase $R_2$ to 0.3 to remedy the COH2 score. 
The result shows that COH1 and COH2 are close to the baseline; at the same time, though, the SCL and LM scores improve. 
%Some examples are shown in Table \ref{table:RL_diff_weight_output_2500}.
%Below we compare this model with external models. %LeeNew: 這句我看不懂 ...

%\FloatBarrier
\begin{table}[h]
\centering
%\begin{tabular}{|c|c|c|c|c|c|c|}
\begin{tabular}{|P{0.8cm}|P{0.8cm}|P{0.8cm}|P{0.8cm}|P{0.8cm}|P{0.8cm}|P{0.8cm}|}
\hline
\multicolumn{3}{|c|}{Input} & COH1 & COH2 & SCL & LM \\ \hline
\multicolumn{3}{|c|}{Seq2seq(baseline)}& -8.6 & 0.664 & 0.33  & -1.574 \\ \hline
$R_{1}$ & $R_{2}$ & $R_{3}$ & \multicolumn{4}{c|}{---} \\ \hhline{|=|=|=|=|=|=|=|}
%0.0 & 0.0 & 1.0 & 11.131 & 0.674 & 0.992 & 0.624\\ \hline
%0.0 & 0.5 & 0.5 & 9.325 & 0.674 & 0.602 & 1.475 \\ \hline
%0.0 & 0.8 & 0.2 & 8.349 & 0.668 & 0.454 & 1.19 \\ \hline
%0.4 & 0.3 & 0.3 & 9.218 & 0.674 & 0.621 & 1.398 \\ \hline

0.0 & 0.0 & 1.0 & -9.518 & 0.589 & 0.992 & -1.471\\ \hline
0.5 & 0.0 & 0.5 & -8.813 & 0.587 & 0.778 & -1.160 \\ \hline
0.8 & 0.0 & 0.2 & -8.419 & 0.588 & 0.658 & -0.967 \\ \hline
0.3 & 0.3 & 0.4 & -8.840 & 0.641 & 0.779 & -0.940 \\ \hline

\end{tabular}
\caption{\it Evaluation of reinforcement learning models with different reward combinations at 2500 iterations on Chinese CECG dataset.}
\label{table:RL_diff_weight_2500}
\end{table}

\subsection{Plug and Play}
VAE is used for sentence generation in the plug and play model, and KL cost annealing and vocabulary truncation are applied to improve VAE performance. 
In this experiment, we first compared the performance between models with and without KL cost annealing. 
These two models truncate vocabulary randomly at a probability of 0.3. 
%In Table~\ref{table:pp_kl_cost}, the model WK loss decreases rapidly; eventually, however, model WK yields higher loss than model K.
In Table~\ref{table:pp_kl_cost}, with KL cost annealing, the KL loss of model increases at first and decreases afterward. 
This observation shows that the model at first decreases the negative log likelihood instead of improving the KL loss. 
This is beneficial to the training process once the model gives high priority to the VAE reconstruction.

%\FloatBarrier
\begin{table}[h]
\centering
\begin{tabular}{|P{1.1cm}|P{1.0cm}|P{0.6cm}|P{0.7cm}|P{0.7cm}|P{0.7cm}|P{0.7cm}|}
\hline
\multicolumn{3}{|c|}{Epochs} & 10000 & 20000 & 30000 & 40000 \\ \hhline{|=|=|=|=|=|=|=|}
\multirow{8}{*}{\begin{tabular}[c]{@{}c@{}}KL cost \\ annealing\end{tabular}} & \multirow{4}{*}{Without} & Total loss & \multirow{2}{*}{28.394} & \multirow{2}{*}{11.163} & \multirow{2}{*}{5.290} & \multirow{2}{*}{3.057} \\ \cline{3-7} 
 &  & KL Loss & \multirow{2}{*}{0.111} & \multirow{2}{*}{0.043} & \multirow{2}{*}{0.021} & \multirow{2}{*}{0.014} \\ \cline{2-7} 
 & \multirow{4}{*}{With} & Total loss & \multirow{2}{*}{23.059} & \multirow{2}{*}{9.186} & \multirow{2}{*}{4.352} & \multirow{2}{*}{2.627} \\ \cline{3-7} 
 &  & KL loss & \multirow{2}{*}{0.132} & \multirow{2}{*}{0.194} & \multirow{2}{*}{0.020} & \multirow{2}{*}{0.037} \\ \hline
\end{tabular}
\caption{\it Loss of plug and play models with or without KL cost annealing under different epochs on Chinese CECG dataset.}
\label{table:pp_kl_cost}
\end{table}

\begin{comment} %LeeNew: 這部分太武斷了 ...
Table~\ref{table:pp_sent_recon1} compares the results of models K and WK. Although both models output grammatical sentences, the output of model WK fails to reconstruct the original sentence. This suggests that as model WK considers less information in the hidden layer encoding than in its own language model decoder, it outputs a sentence close to but not identical to the original sentence.
%\FloatBarrier
\begin{table}[h]
\centering
\begin{tabular}{|P{2.1cm}|P{1.2cm}|p{3.9cm}|}
\hline
\multicolumn{2}{|c|}{\multirow{2}{*}{Original input}} & 這是俄羅斯美食啊\\
\multicolumn{2}{|c|}{} & This is Russian cuisine\\
\hhline{|=|=|=|}
\multirow{4}{*}{KL cost annealing} &  \multirow{2}{*}{Used} & 這是日本美食啊\\
 & & This is Japanese cuisine\\
\cline{2-3} 
 & \multirow{2}{*}{Not used} & 這是俄羅斯美食啊\\
 & & This is Russian cuisine\\
\hline
\end{tabular}
\caption{\it Sentence reconstruction of plug and play models with or without KL cost annealing on Chinese CECG dataset.}
\label{table:pp_sent_recon1}
\end{table}
\end{comment}

Table~\ref{table:pp_eval_vocab_truncate} compares three difference truncation probabilities: 0.0, 0.3, and 0.7 (0.0 corresponds to no vocabulary truncation). 
The best is 0.3, and the worst is 0.7. 
This shows that a small proportion of truncation helps VAE to depend less on its own language model; a high proportion hinders training.

%\FloatBarrier
\begin{table}[h]
\centering
\begin{tabular}{|P{1.7cm}|P{0.4cm}|P{0.8cm}|P{0.6cm}|P{0.7cm}|P{0.6cm}|P{0.6cm}|}
\hline
\multicolumn{3}{|c|}{Input} & 10000 & 20000 & 30000 & 40000 \\ \hhline{|=|=|=|=|=|=|=|}
\multirow{12}{*}{\begin{tabular}[c]{@{}c@{}}Vocabulary\\ truncation \%\end{tabular}} & \multirow{4}{*}{0.0} & Total & \multirow{2}{*}{24.622} & \multirow{2}{*}{11.003} & \multirow{2}{*}{4.474} & \multirow{2}{*}{2.675} \\
 & & loss & & & & \\
\cline{3-7} 
 & & KL & \multirow{2}{*}{0.176} & \multirow{2}{*}{1.568} & \multirow{2}{*}{0.056} & \multirow{2}{*}{0.092} \\ 
 & & loss & & & & \\
\cline{2-7} 
 & \multirow{4}{*}{0.3} & Total & \multirow{2}{*}{23.059} & \multirow{2}{*}{9.186} & \multirow{2}{*}{4.352} & \multirow{2}{*}{2.627} \\
 & & loss & & & & \\
\cline{3-7} 
 & & KL & \multirow{2}{*}{0.132} & \multirow{2}{*}{0.194} & \multirow{2}{*}{0.020} & \multirow{2}{*}{0.037} \\
 & & loss & & & & \\
\cline{2-7} 
 & \multirow{4}{*}{0.7} & Total loss & \multirow{2}{*}{25.361} & \multirow{2}{*}{10.140} & \multirow{2}{*}{4.962} & \multirow{2}{*}{2.830} \\
\cline{3-7} 
 & & KL & \multirow{2}{*}{0.164} & \multirow{2}{*}{0.084} & \multirow{2}{*}{0.036} & \multirow{2}{*}{0.014} \\
 & & loss & & & & \\ 
 \hline
\end{tabular}
\caption{\it Plug and play model loss with different vocabulary truncation proportions under different epochs on Chinese CECG dataset.}
\label{table:pp_eval_vocab_truncate}
\end{table}

In Table~\ref{table:pp_sent_recon2}, we see that the 0.3 output successfully reconstructs the original input. 
The 0.0 output also seems reasonable, although it fails to reconstruct the original sentence: it replaces ``禮貌'' (``courtesy'') with ``動作'' (``action''), probably because it depends more on the decoder language model. 
With 0.7 truncation, the model outputs an incorrect sentence, replacing ``提高'' (``improve'') with ``以'' (``by''); this could be because the many unknown words hinder the model from learning correct sentence grammar. 
The model with KL cost annealing and with a vocabulary truncation of 0.3 outperforms all others, so we will use this setting in the following experiments. 
%model and hyperparameter set.  % AMH: not sure what this means

%\FloatBarrier
\begin{table}[h]
\centering
\begin{tabular}{|P{1.6cm}|P{0.7cm}|p{4.9cm}|}
\hline
\multicolumn{2}{|c|}{\multirow{2}{*}{Original input}} & 禮貌能提高一個的素質\\
\multicolumn{2}{|c|}{} & Courtesy can improve quality\\
\hhline{|=|=|=|}
\multirow{6}{*}{\begin{tabular}[c]{@{}c@{}}Vocabulary\\ truncation \%\end{tabular}} & \multirow{2}{*}{0.0} & 動作能提高一個的素質\\
 & & Action can improve quality\\
\cline{2-3}
 & \multirow{2}{*}{0.3} & 禮貌能提高一個的素質\\
 & & Courtesy can improve quality\\
\cline{2-3} 
 & \multirow{2}{*}{0.7} & 禮貌能以一個的素質\\
 & & Courtesy can be by quality\\
\hline
\end{tabular}
\caption{\it Sentence reconstruction of plug and play models with different vocabulary truncation proportions on Chinese CECG dataset.}
\label{table:pp_sent_recon2}
\end{table}

Tables~\ref{table:pp_gen} apply sentiment gradient ascent and descent respectively. 
%In Table~\ref{table:pp_gen1}, we note the short sequence length of the output before sentiment transfer. 
Both positive and negative sentiment transfers generate corresponding sentiment outputs. However, most sentiment transfers replace words with strong sentiment bias. 
For example, in Table~\ref{table:pp_gen}, ``哈哈'' (``haha'') is added after a positive transfer (sentiment gradient ascent), and ``淘汰'' (``eliminate'') is added after a negative transfer (sentiment gradient descent); although each of these words is a strong indicator of sentiment style, it leads to incorrect grammar.

\begin{table*}[htbp]
  \centering
    \captionsetup{justification=centering}
    \begin{tabular}{|c|c|l|l|}
    \hline
    \multicolumn{2}{|c|}{\multirow{2}{*}{Original input}} & 果然昇仙了大霧... & 終於搞完了，接待真不是人幹的活[怒罵]\\
    \multicolumn{2}{|c|}{} & Sure enough, it was a big fog... & Finally finished~-- reception sucks! [roaring]\\
    \hline
    \multicolumn{2}{|c|}{\multirow{2}{*}{Original output}} & 慢走。。 & 你們這接待搞得人都徹底消失了。\\
    \multicolumn{2}{|c|}{} & Walk yourself out.. & Your ``reception'' has driven everyone away.\\
    \hhline{|=|=|=|=|}
    \multirow{4}{*}{\begin{tabular}[c]{@{}c@{}}Sentiment\\ score\end{tabular}} & \multirow{2}{*}{\begin{tabular}[c]{@{}c@{}}Gradient\\ ascent\end{tabular}} & 期待豁。 & 哈哈規定地方搞得人都徹底消失了。\\
     & & Looking forward to it. & Ha ha, stipulating the location has driven everyone away.\\
    \cline{2-4} 
     & \multirow{2}{*}{\begin{tabular}[c]{@{}c@{}}Gradient\\ descent\end{tabular}} & 安息。。 &  你們這搞找得人都成淘汰了\\
     & & Rest in peace.. & You've made everyone to eliminated.\\
    \hline
    \end{tabular}
  \caption{\it Sentences generated by plug and play models on Chinese CECG dataset.}
  \label{table:pp_gen}
\end{table*}

%Word embedding of 300 dimensions is used in training phase, and the max length of a sentence is 15.%

\subsection{CycleGAN}
When training the CycleGAN model, heterogeneous input styles are preferred; thus the data used here is re-marked and selected by the sentiment classifier, leaving only those sentences with strong style bias. 
In the CycleGAN model training phase, training alternates between the generator and discriminator. Once one of the agents (generator or discriminator) is trained, the other's parameters are fixed and used for inference. If one is better than the other, a training imbalance occurs, and the process is no longer ``adversarial''. 
Since the discriminator is easier to train in most cases, the discriminator is stronger than the generator generally, leading to an awkward generator. We evaluate two kinds of generators: generators G and F, created after the generator and the discriminator are trained alternatively using different training epochs. 
Furthermore, we also tried to add Identity Loss (short for ID loss), which was used to help to keep the content after transferring, in total loss during training the generator. 
The result is illustrated in Tables~\ref{table:cyclegan_diff_epochs10000} and \ref{table:cyclegan_diff_epochs100000}: the training epoch ratios of the generator and discriminator are 1:1, 1:3, and 3:1 after 10000 iterations and 100000 iterations respectively.

%The 1:1 result is most stable: generators F and G are both best at 100000 iterations even though generator F is worse at 10000 iterations, meaning the generator and the discriminator are trained and optimized together. 
The 1:1 result is most stable.
For the 1:3 model, although it achieves good performance at 10000 iterations, the model's generator and discriminator both collapse at 100000 iterations; this may be because the generator is unable to deceive the robust discriminator after several iterations of training; repeated training in this case leads to collapse.  % AMH: check
Although the 3:1 model is relatively stable, its generator is not as good as that of the 1:1 model. 
%This may be because the discriminator was trained too well in the early stages. 
Moreover, the experiment also shows if there is an improvement when the ID loss  is added: the result shows no significant difference from when only generator loss is used as the total loss. 
%This might be attributable to sentence generation tasks being far easier than image generation tasks.

%\FloatBarrier
\begin{table}[h]
\centering
\begin{tabular}{|>{\centering\arraybackslash}P{0.8cm}|c|c|c|c|c|c|}\hline
\multicolumn{3}{|c|}{Epochs} & \multicolumn{1}{c|}{COH1} & \multicolumn{1}{c|}{COH2} & \multicolumn{1}{c|}{SCL} & \multicolumn{1}{c|}{LM} \\ 
\hline
\begin{tabular}[c]{@{}c@{}}Gener-\\ator\\type\end{tabular} & \multicolumn{1}{c|}{\begin{tabular}[c]{@{}c@{}}Generator\\ epochs\end{tabular}} & \multicolumn{1}{c|}{\begin{tabular}[c]{@{}c@{}}Discri-\\minator\\ epochs\end{tabular}} & \multicolumn{4}{c|}{---} \\ 
\hhline{|=|=|=|=|=|=|=|}
\multirow{4}{*}{\shortstack[c]{G : \\neg\\to\\pos}} & 1 & 1 & -9.071 & 0.665 & 0.683 & -4.171 \\ 
\cline{2-7} 
 & 3 & 1 & -9.145 & 0.668 & 0.602 & -4.326 \\
\cline{2-7} 
 & 1 & 3 & -9.154 & 0.675 & 0.679 & -4.237 \\
\cline{2-7} 
 & 1(ID loss) & 1 & -9.064 & 0.667 & 0.659 & -4.193 \\
\hline
\multirow{4}{*}{\shortstack[c]{F : \\pos\\to\\neg}} & 1 & 1 & -9.215 & 0.666 & 0.279 & -4.427 \\ \cline{2-7} 
 & 3 & 1 & -9.229 & 0.667 & 0.254 & -4.369 \\ \cline{2-7} 
 & 1 & 3 & -9.244 & 0.664 & 0.203 & -4.285 \\ \cline{2-7} 
 & 1(ID loss) & 1 & -9.22 & 0.664 & 0.25 & -4.426 \\
\hline
\end{tabular}
\caption{\it CycleGAN model under different training epoch combinations at a total of 10000 epochs on the Chinese CECG dataset.}
\label{table:cyclegan_diff_epochs10000}
\end{table}

%\FloatBarrier
\begin{table}[h]
\centering
%\begin{tabular}{|>{\centering\arraybackslash}P{1.0cm}|>{\centering\arraybackslash}P{1.1cm}|>{\centering\arraybackslash}P{0.5cm}|c|c|c|c|}
\begin{tabular}{|>{\centering\arraybackslash}P{0.8cm}|c|c|c|c|c|c|}
\hline
\multicolumn{3}{|c|}{Epochs} & \multicolumn{1}{c|}{COH1} & \multicolumn{1}{c|}{COH2} & \multicolumn{1}{c|}{SCL} & \multicolumn{1}{c|}{LM} \\ 
\hline
\begin{tabular}[c]{@{}c@{}}Gener-\\ator\\type\end{tabular} & \multicolumn{1}{c|}{\begin{tabular}[c]{@{}c@{}}Generator\\ epochs\end{tabular}} & \multicolumn{1}{c|}{\begin{tabular}[c]{@{}c@{}}Discri-\\minator\\ epochs\end{tabular}} & \multicolumn{4}{c|}{---} \\ 
\hhline{|=|=|=|=|=|=|=|}
\multirow{4}{*}{\shortstack[c]{G : \\neg\\to\\pos}} & 1 & 1 & -9.206 & 0.667 & 0.682 & -4.13 \\ 
\cline{2-7} 
 & 3 & 1 & -9.2 & 0.669 & 0.654 & -4.227 \\
\cline{2-7} 
 & 1 & 3 & -17.341 & 0.536 & 0.013 & -8.731 \\ \cline{2-7} 
 & 1(ID loss) & 1 & -9.197 & 0.667 & 0.664 & -4.121 \\
\hline
\multirow{4}{*}{\shortstack[c]{F : \\pos\\to\\neg}} & 1 & 1 & -9.192 & 0.658 & 0.201 & -4.662 \\ \cline{2-7} 
 & 3 & 1 & -9.203 & 0.666 & 0.222 & -4.418 \\ \cline{2-7} 
 & 1 & 3 & -17.341 & 0.536 & 0.013 & -8.731 \\ \cline{2-7} 
 & 1(ID loss) & 1 & -9.167 & 0.665 & 0.213 & -4.422 \\
\hline
\end{tabular}
\caption{\it CycleGAN model under different training epoch combinations at a total of 100000 epochs on the Chinese CECG dataset.}
\label{table:cyclegan_diff_epochs100000}
\end{table}

The 1:1 model with 100000 training iterations is the best model; there will be a comparison between this model and other trained models later. 
Tables~\ref{table:cyclegan_senttrans} contain examples of sentences transformed by generators G and F. 
In the left table of Table~\ref{table:cyclegan_senttrans}, as the un-transformed output is already a positive response, generator G has little effect on the result; generator F, however, produces an identifiable result with its clearly negative sentence. In the right table of Table~\ref{table:cyclegan_senttrans}, the original output is a negative response; generator G substitutes ``討厭'' (``hate'') for ``喜歡'' (``like''), making a more positive sentence. Generator F makes little modification to the sentence.

\begin{table*}[htbp]
  \small
  \centering
    \captionsetup{justification=centering}
    \begin{tabular}{|c|p{5.2cm}|l|}
    \hline
    \multirow{2}{*}{Original input} & 再見廈門。再見朋友們。。 & 老天爺為什麼要發明洗頭這項運動[淚]\\
     & Goodbye Xiamen. Goodbye friends. & Why did God invent shampooing? [sob]\\
    \hline
    \multirow{3}{*}{Original output} &
    廈門人民隨時歡迎你歸來！[酷] & 我最討厭的就是洗頭\\
     & The people of Xiamen welcome you back any time! [cool] & What I hate most is shampooing \\
    \hhline{|=|=|=|}
    \multirow{3}{*}{Generator G} & 廈門人民隨時歡迎你歸來！[酷酷] & 我最喜歡的就是洗頭\\
     & The people of Xiamen welcome you back any time! [cool cool] & What I hate most is shampooing\\
    \hline
    \multirow{3}{*}{Generator F} &
    廈門人民隨時怕你歸來！[抓狂] & 我真討厭的就是洗頭\\
     & The people of Xiamen will always fear your return! [crazy] & What I really hate is shampooing\\
    \hline
    \end{tabular}
  \caption{\it Senteces generated by the CycleGAN model on Chinese CECG dataset.}
  \label{table:cyclegan_senttrans}
\end{table*}
%我其實很困惑，為什麼每個 model 選的 example 都不一樣啊 ...... Ho: 我猜是好的Example不好找...

\section{Comparison of All Models}\label{sec:comparison}

\subsection{Chinese CECG Dataset}

Table~\ref{table:scores_chinese} evaluates the four models using four metrics. 
The results better than the baselines are in blue, and the ranking of each method is also in the table. 
The persona-based model (using sentiment score equal to 1.0 as input during inference) yields the highest SCL.
The RL model yields the best performance on COH1 and LM. %all metrics, except sentiment score. 
 %because it uses the sentiment score as input directly during training, and uses 1.0 as input during inference; thus its performance is reasonable. %這個推論合理嗎?
The plug and play and CycleGAN models transform the original output sentences directly, resulting in similar SCL scores, but worse than persona-based model and RL model.
CycleGAN has better LM score than plug and play, which shows that CycleGAN does better with respect to grammar. 
%grammar and semantic meaning, but gets worse result on sentiment; It is because the reinforcement learning model considers not only  sentiment but also semantic meaning and dialogue coherence as rewards for training that diversifies outputs, but might sacrifice sentiment performance to some extent. Instead of modifying parameters of the original model like personal-based and reinforcement learning model, plug and play and CycleGAN model transform original output sentences directly that the scores of the metrics are close to each other, yet the CycleGAN gets better result on grammar.

%\FloatBarrier
\begin{table}[h]
\centering
\begin{tabular}{|P{2.0cm}|P{1.4cm}|P{1.0cm}|P{1.0cm}|P{1.2cm}|}
\hline
\multirow{2}{*}{\backslashbox{Model}{Metrics}} & \multicolumn{2}{c|}{Semantic Coh.} & 
Sent. & Lang. \\
\cline{2-5}
& COH1 & COH2 & SCL & LM\\
\hhline{|=|=|=|=|=|}
Seq2seq (baseline)     & $-8.600$ & $0.664$                  & $0.330$ & $-1.574$\\
\hhline{|=|=|=|=|=|}
Persona-based          &\ding{195} $-9.338$ &\ding{194} $0.639$                  &\ding{192} \textcolor{blue}{$0.925$} &\ding{193} $-2.178$\\
\hline 
Reinforcement L.       &\ding{192} $-8.840$ &\ding{193} $0.641$                  &\ding{193} \textcolor{blue}{$0.779$} &\ding{192} \textcolor{blue}{$-0.940$} \\
\hline
Plug and Play          &\ding{193} $-9.158$ &\ding{195} $0.602$                  &\ding{194} \textcolor{blue}{$0.682$} &\ding{195} $-4.930$\\
\hline
CycleGAN               &\ding{194} $-9.206$ &\ding{192} \textcolor{blue}{$0.667$} &\ding{194} \textcolor{blue}{$0.682$} &\ding{194} $-4.130$\\
\hline
\end{tabular}
\caption{\it Evaluation on Chinese CECG dataset. COH1, COH2, SCL, and LM stand for semantic coherence 1, semantic coherence 2, sentiment classifier score, and language model score, respectively. The results better than the baselines are in blue, and the ranking of each method is also in the tables. }
\label{table:scores_chinese}
\end{table}

\subsection{Chinese PTT Dataset}\label{section:PTT}

In this section, we switched to the PTT dataset to demonstrate the performance of the above four models. To train a sentiment classifier, we also tried CNN, GRU-last, and GRU-avg on different word segmentation methods at first. However, as the dirty PTT dataset resulted in poor sentiment classifier performance,  we improved the sentiment classifier using the following three steps.
First, we compared the performance between CNN, GRU-last, and GRU-avg, and chose GRU-last because of its out-performance (see Table~\ref{table:ptt_sent_classifier}). 
Second, we picked the more credible data (higher absolute scores) by GRU-last, shrinking the data from 1,402,303 to 1,149,000.
Finally, we trained a new sentiment classifier on the remaining data using GRU-last-2. This final sentiment classifier was then used in the four models. GRU-last-2 here indicates the same architecture as GRU-last but not the same model.

%\FloatBarrier
\begin{table}[h]
\centering
\footnotesize
\begin{tabular}{|P{0.6cm}|P{0.6cm}|P{0.6cm}|P{0.6cm}|P{0.7cm}|P{0.7cm}|P{0.7cm}|P{0.6cm}|}
\hline
Segm. & \multicolumn{4}{c|}{Word-based} & \multicolumn{3}{c|}{Character-based}\\
\hline

\multirow{2}{*}{Struct.} & \multirow{2}{*}{CNN} & GRU-last & GRU-avg & GRU-last-2 &  \multirow{2}{*}{CNN} & GRU-last & GRU-avg \\
\hhline{|=|=|=|=|=|=|=|=|}
Acc&$\textnormal0.803$&$0.843$&$0.842$&$0.994$&$0.813$&$0.848$&$0.846$\\
\hline 
AUC&$\textnormal0.5$&$0.849$&$0.848$&$0.998$&$0.5$&$0.831$&$0.824$\\
\hline
\end{tabular}
\caption{\it Sentiment classifiers of different NN architectures on PTT dataset.}
\label{table:ptt_sent_classifier}
\end{table}

The evaluation results on the PTT dataset are shown in Table~\ref{table:scores_ptt}.
SCL scores are all high in these four models.
It shows that all the four approaches can successfully generate positive responses, but persona-based and RL models are better than plug and play and Cycle GAN.  
Although the SCL score of persona-based model is the highest, it obtained the worst COH1 and COH2, which shows that its generated responses are not coherent with respect to the inputs.
Similar to that on the CECG dataset, the RL model yields the best results in terms of COH1 and LM. 

%except that it obtains worse COH2 score than plug and play model and CycleGAN, and its SCL score is slightly lower than persona-based model.

%better performance on all the evaluation metrics, except coherence score 2 (worse the plug and play model).   
%are quite similar to that on the CECG dataset; the reinforcement learning model yields better performance on all the evaluation metrics, except coherence score 2 (worse the plug and play model).   
%The persona-based model performs poorly as there are too many unknown word in the output due to the PTT dataset's many special words. 

%\FloatBarrier
\begin{table}[h]
\centering
\begin{tabular}{|P{2.0cm}|P{1.4cm}|P{1.0cm}|P{1.0cm}|P{1.4cm}|}
\hline
\multirow{2}{*}{\backslashbox{Model}{metrics}} & \multicolumn{2}{c|}{Semantic coh.} & 
Sent. & Lang. \\
\cline{2-5}
& COH1 & COH2 & SCL & LM\\
\hhline{|=|=|=|=|=|}
Seq2seq(baseline)    & $-9.783$ & $0.553$ & $0.508$ & $-1.778$\\
\hhline{|=|=|=|=|=|}
Persona-based        &\ding{195} $-11.283$ &\ding{195}             $0.479$                &\ding{192} \textcolor{blue}{$0.995$} &\ding{193} $-2.249$\\
\hline 
Reinforcement L.     &\ding{192} \textcolor{blue}{$-9.527$} &\ding{194} $0.511$            &\ding{193} \textcolor{blue}{$0.991$} &\ding{192} \textcolor{blue}{$-1.577$}\\
\hline
Plug and play        &\ding{193} \textcolor{blue}{$-9.700$} &\ding{193} \textcolor{blue}{$0.560$} &\ding{195} \textcolor{blue}{$0.974$} &\ding{195} $-2.793$\\
\hline
CycleGAN             &\ding{194} $-9.811$                  &\ding{192} \textcolor{blue}{$0.573$} &\ding{194} \textcolor{blue}{$0.984$} &\ding{194} $-2.678$\\
\hline
\end{tabular}
\caption{\it Model evaluation on PTT dataset.}
\label{table:scores_ptt}
\end{table}

\subsection{English Corpus} 
%Some examples are shown on the following link: \url{goo.gl/X1PZLM}.

The results are listed in Table~\ref{table:scores_english}. 
For the persona-based model, its SCL score is the highest; however, its COH1, COH2, and LM are the lowest. 
The RL model performed better than all other models in three out of the four metrics: COH1, COH2, and LM, but not the SCL score.
The SCL scores of plug and play model and CylcleGAN are worse than persona-based and RL models. 
%The COH1 and COH2 scores of plug and play model were both lower than that of most others. 
%The COH1 and COH2 scores of cycleGAN were not too far from the seq2seq baseline.

%\FloatBarrier
\begin{table}[htb]
\centering
\begin{tabular}{|P{2.0cm}|P{1.4cm}|P{1.0cm}|P{1.0cm}|P{1.4cm}|}
\hline
\multirow{2}{*}{\backslashbox{Model}{metrics}} & \multicolumn{2}{c|}{Semantic coh.} & 
Sent. & Lang. \\
\cline{2-5}
& COH1 & COH2 & SCL & LM\\
\hhline{|=|=|=|=|=|}
Seq2seq(baseline)          & $-0.755$ & $0.762$                   & $0.543$ & $-1.465$\\
\hhline{|=|=|=|=|=|}
Persona-based              &\ding{195} $-1.961$ &\ding{195} $0.710$                   &\ding{192} \textcolor{blue}{$0.870$} &\ding{195} $-2.169$\\
\hline 
Reinforcement L.           &\ding{192} $-0.839$ &\ding{192} \textcolor{blue}{$0.792$} &\ding{193} \textcolor{blue}{$0.777$} &\ding{192} $-1.556$\\
\hline
Plug and play              &\ding{194} $-1.364$ &\ding{194} $0.759$                   &\ding{194} \textcolor{blue}{$0.697$} &\ding{194} $-1.671$\\
\hline
CycleGAN                   &\ding{193} $-0.979$ &\ding{193} \textcolor{blue}{$0.764$} &\ding{195} \textcolor{blue}{$0.695$} &\ding{193} $-1.562$\\
\hline
\end{tabular}
\caption{\it Evaluation of models on English dataset.}
\label{table:scores_english}
\end{table}

\subsection{Examples}

Tables~\ref{table:examples} list outputs of the four models on Chinese CECG dataset. 
%Tables~\ref{table:example1_ptt} and \ref{table:example2_ptt} are examples of generated outputs of Chinese PTT dataset. 
The persona-based and RL models usually output reasonable and positive sentences. 
From the data we also observed both the persona-based and RL models made complicated changes to output sentences; this was rarely seen in other models.
The plug and play model tends to generate collapsed sentences. 
The CycleGAN model typically adjusts sentences by replacing negative words with positive ones; this is not always a significant transformation.

\begin{table*}[htbp]
  \small
  \centering
  \captionsetup{justification=centering}
    \begin{tabular}{|c|l|l|}
    \hline
    \multirow{4}{*}{Input} &
    想到一個還不錯的點子，對自己的智商有了信心。 & 要吃水煮魚了哇賽好開心\\
     & Come out a good idea and is confident in my IQ now. & I'm going to eat boiled fish, wow I’m so excited\\
    \hline
    \multirow{2}{*}{Original output} & 也許是錯覺[淚] & 哪家好吃\\
     & Could be an illusion [sob] & Which restaurant do you recommend?\\
    \hhline{|=|=|=|}
    \multirow{2}{*}{Persona-based} & 哈哈，你是不是覺得你的智商是很不錯的 & 哈哈，我也是\\
     & Haha, you think you have a high IQ? & Haha, me too\\
    \hline 
    \multirow{2}{*}{Reinforcement L.} & 贊同。 & 哈哈，我也想去吃呢。\\
     & I agree. & Haha, I want to get some too.\\
    \hline 
    \multirow{2}{*}{Plug and Play} & 波波那。）瞬間 & 好好吃？\\
     & Bobona. ) suddenly & Is it good?\\
    \hline 
    \multirow{2}{*}{CycleGAN} & 也許是錯覺[哈哈] & 哪家好吃\\
     & Could be an illusion [Haha] & Which restaurant do you recommend?\\
    \hline 
    \end{tabular}
  \caption{\it Sentences generated by four models on Chinese CECG dataset.}
  \label{table:examples}
\end{table*}

\begin{comment}
%\FloatBarrier
\begin{table}[h]
\centering
\begin{tabular}{|P{2.3cm}|p{5.4cm}|}
\hline
\multirow{2}{*}{Input} & 我放生了公主女友.\\
 & I dumped my princess syndrome girl friend.\\
\hline
\multirow{2}{*}{Original Output} & 噓你放生\\
 & Boo because of your dumping\\
\hhline{|=|=|}
\multirow{2}{*}{Personal-based} & 你是不是被她洗腦的感覺\\
 & It seems like you were brainwashed by her\\
\hline 
\multirow{2}{*}{Reinforcement L.} & 恭喜你！\\
 & Congratulations!\\
\hline 
\multirow{2}{*}{Plug and Play} & 噓你放生\\
 & Boo because of your dumping\\
\hline 
\multirow{2}{*}{CycleGAN} & 推放生\\
 & like your dumping\\
\hline 
\end{tabular}
\caption{\it Sentences generated by four models on PTT dataset (Example 1).}
\label{table:example1_ptt}
\end{table}

%\FloatBarrier
\begin{table}[htb]
\centering
\begin{tabular}{|P{2.3cm}|p{5.4cm}|}
\hline
\multirow{2}{*}{Input} & 感動與心動？\\
 & Touched and moved ?\\
\hline
\multirow{2}{*}{Original Output} & 朝聖先噓\\
 & Pilgrimage and boo first\\
\hhline{|=|=|}
\multirow{2}{*}{Personal-based} & 朝聖推\\
 & Pilgrimage\\
\hline 
\multirow{2}{*}{Reinforcement L.} & 推！\\
 & Like!\\
\hline 
\multirow{2}{*}{Plug and Play} & 推優文\\
& Like the paragraph\\
\hline 
\multirow{2}{*}{CycleGAN} & 朝聖推\\
 & Pilgrimage\\
\hline 
\end{tabular}
\caption{\it Sentences generated by four models on PTT dataset (Example 2).}
\label{table:example2_ptt}
\end{table}
\end{comment}

\section{Human Evaluation}\label{sec:human}
%\vspace{-1mm}

We performed a subjective human evaluation against sentences generated by five models each on three corpora (CECG, PTT, and English), for a total of 15 results with 30 subjects, all of whom were graduate students.
They were asked to answer three questions about the output sentences: (1) Coherence: Is the output sentence a good response to the input? (2) Sentiment: Is the output sentence positive? (3) Grammar: Is the output sentence grammatically correct?
They were asked to give scores ranging from $0$ to $5$, based on a few reference examples with given scores $1$, $3$, $5$ to scale the scores.
The average results (normalized to the range from $0$ to $1$) of three models are listed in Tables~\ref{table:human_CECG}, \ref{table:human_ptt}, and \ref{table:human_english}.
The results better than the baselines are in blue, and the ranking of each method is also in the tables. 

For coherence, all the approaches are worse than the baselines, except RL models on  Chinese PTT dataset.
Except for the second rank on the English corpus, the RL models  worked the best among the four approaches in terms of coherence.
For sentiment, all the models can successfully modify the sentiment of responses, except plug and play models, which performed the worst on every corpus. 
For grammar, the RL models yielded the best performance on each corpus. 

%Since the subjective human evaluation questions are parallel to the objective machine evaluation scores, we calculate the Pearson correlation coefficients $\rho$ between Coherence, Sentiment and Grammar scores in Table \ref{table:human_english} with respect to COH1, SCL, and LM scores in Table \ref{table:scores_english}.
%The results are $0.728$, $0.885$ and $0.543$ respectively.
%This shows the machine evaluation metrics used here were well correlated to the human evaluation. 

%\FloatBarrier
\begin{table}[h]
\centering
\begin{tabular}{|P{2.8cm}|P{1.35cm}|P{1.35cm}|P{1.35cm}|}
\hline
& Coherence & Sentiment & Grammar \\
\hhline{|=|=|=|=|}
Seq2seq(baseline)&$\textnormal 0.782$&$0.583$&$0.818$\\
\hhline{|=|=|=|=|}
Persona-based    &\ding{193} $0.670$   &\ding{194} \textcolor{blue}{$0.647$}   &\ding{193} $0.734$\\
\hline 
Reinforcement L. &\ding{192} $0.777$   &\ding{192} \textcolor{blue}{$0.734$}   &\ding{192} \textcolor{blue}{$0.822$}\\
\hline
Plug and play    &\ding{195} $0.534$   &\ding{195} $0.522$                     &\ding{195} $0.500$\\
\hline
CycleGAN         &\ding{194} $0.645$   &\ding{193} \textcolor{blue}{$0.667$}   &\ding{194} $0.680$\\
\hline
\end{tabular}
\caption{Human evaluation scores on Chinese CECG dataset for coherence, sentiment, and grammar. Average scores normalized to $[0,1]$. The results better than the baselines are in blue, and the ranking of each method is also in the table. }
\label{table:human_CECG}
\end{table}

%\FloatBarrier
\begin{table}[h]
\centering
\begin{tabular}{|P{2.8cm}|P{1.35cm}|P{1.35cm}|P{1.35cm}|}
\hline
& Coherence & Sentiment & Grammar \\
\hhline{|=|=|=|=|}
Seq2seq(baseline)&$\textnormal 0.704$&$0.637$&$0.531$\\
\hhline{|=|=|=|=|}
Persona-based    &\ding{194} $0.631$                       &\ding{193} \textcolor{blue}{$0.648$}     &\ding{193} \textcolor{blue}{$0.697$}\\
\hline 
Reinforcement L. &\ding{192} \textcolor{blue}{$0.712$}     &\ding{194} \textcolor{blue}{$0.645$}     &\ding{192} \textcolor{blue}{$0.746$}\\
\hline
Plug and play    &\ding{195} $0.528$                       &\ding{195} $0.574$                       &\ding{195} \textcolor{blue}{$0.581$}\\
\hline
CycleGAN         &\ding{193} $0.641$                       &\ding{192} \textcolor{blue}{$0.686$}     &\ding{194} \textcolor{blue}{$0.689$}\\
\hline
\end{tabular}
\caption{Human evaluation scores on Chinese PTT dataset for coherence, sentiment, and grammar. Average scores normalized to $[0,1]$.}
\label{table:human_ptt}
\end{table}

%\FloatBarrier
\begin{table}[h]
\centering
\begin{tabular}{|P{2.8cm}|P{1.35cm}|P{1.35cm}|P{1.35cm}|}
\hline
& Coherence & Sentiment & Grammar \\
\hhline{|=|=|=|=|}
Seq2seq(baseline)&$\textnormal 0.548$&$0.161$&$0.999$\\
\hhline{|=|=|=|=|}
Persona-based    &\ding{194} $0.235$      &\ding{192}  \textcolor{blue}{$0.705$}       &\ding{194}  $0.746$\\
\hline 
Reinforcement L. &\ding{193}  $0.346$     &\ding{193}  \textcolor{blue}{$0.698$}       &\ding{192}  $0.925$\\
\hline
Plug and play    &\ding{195}  $0.150$     &\ding{195}  \textcolor{blue}{$0.483$}       &\ding{195}  $0.430$\\
\hline
CycleGAN         &\ding{192}  $0.435$     &\ding{194}  \textcolor{blue}{$0.627$}       &\ding{193}  $0.912$\\
\hline
\end{tabular}
\caption{Human evaluation scores on English dataset for coherence, sentiment, and grammar. Average scores normalized to $[0,1]$.}
\label{table:human_english}
\end{table}

\section{Discussion}
The persona-based model and RL model can both generate the responses largely different from the original seq2seq model. %transform the output response.
For the persona-based model, its sentiment score was the highest on all the datasets in terms of both machine and human evaluation (the only exception is the human evaluation on Chinese CECG).
However, its coherence and grammar were both worse than the RL model in terms of both machine and human evaluation.  
This shows that although persona-based model can successfully generate very positive responses, the coherence and the grammar of the responses are poor.
It tries to output sentences that carry the correct sentiment, but not necessarily relevant given the input.
The RL model is generally the best model of the three corpora from all aspects. 
This is because the reward $R_1$ and $R_2$ in Eqs.~(\ref{equ:lm}) and (\ref{equ:ch}) were in parallel with coherence, and $R_1$ in Eq.~(\ref{equ:lm}) also takes into account word ordering which leads to correct grammar.
Its sentiment score was also high (although not as high as the persona-based model) because its reward $R_3$ is also in parallel with the sentiment, which yielded positive output.
One may argue that the RL model overfits the machine-evaluation metrics since it learns to optimize those metrics.
This issue is addressed in Section~\ref{sec:human} through human evaluation.
Humans also consider that the RL model has reasonable responses with correct grammar.

The plug and play model and CycleGAN only transform the output responses of an off-the-shelf seq2seq model.
The plug and play model attempted to modify the latent code of the sentences. 
As the sentiment classifier primarily considered sentiment without really encoding sentence grammar, when maximizing the sentiment classifier's output, it sometimes transforms the original output of the seq2seq model into collapsed sentences.  
For CycleGAN, since the two translators directly outputted word embeddings carrying both sentiment and semantics, the models found mappings between words like ``bad" to ``good'', ``sorry'' to ``thank'', ``can't'' to ``can''. 
However, this entailed only changes or deletions of specific words and not complex modifications of whole sentences.
Since it only changes a few words of the original responses, its responses are not far from the original ones.
This explains why the grammar of plug and play is worse than CycleGAN in terms of both machine and human evaluation in all cases. 
When it comes to sentiment and coherence, humans always consider CycleGAN is better than plug and play, whereas their performances on COH1, COH2, and SCL are comparable. 
Because it is difficult for humans to read the sentences with poor grammar,  humans usually consider that the sentiments of the collapsed sentences are incorrect, and they are less coherent with the inputs.

\section{Conclusion} \label{sec:conclusion}

In this paper, we attempted to adjust the sentiment of the chatbot response given the input. 
We investigated four different models for the tasks; All of the models are based on the conventional seq2seq model.
The performances of the four models in terms of machine-evaluated metrics and human evaluation are reported.
%After careful evaluation and analysis for the four proposed models on different aspects, we found:
The persona-based model and RL models, which alter the original seq2seq model parameters, yield good results.
%Apart from the PTT dataset results in Table~\ref{table:scores_ptt}, Tables~\ref{table:scores_chinese} and \ref{table:scores_english} show consistent results between Chinese and English experiments. The peculiar result of Table~\ref{table:scores_ptt} may be due to its poor data quality.
The persona-based model is good at exporting sentences of high sentiment score that might be suitable for cases when a chatbot only needs to reply with simple sentences that carry strong sentiment. 
The RL model generates high quality sentences, which is likely to prolong conversations. 
On the other hand, if there is already a running, functional chatbot, and the only thing to do is to transfer its sentiment (or style), then the CycleGAN model might be a better choice than plug and play.
As the CycleGAN model primarily performs word mappings on the original response, the output sentence quality is more or less preserved.
The plug and play model currently yields poor performance, probably because it is difficult to modify the latent code of a sentence while preserving its semantics and sentence quality.

\begin{comment}
\begin{enumerate}
\item In Chinese, the model of personal-based and of reinforcement learning, which alter the original pre-trained MLE model parameters, get good result. The personal-based model is good at exporting sentences of high sentiment score that might be suitable for cases like a chatbot only need to reply simple and strong sentiment sentences. The reinforcement model  is capable to export more informative sentences, which might be more likely to let conversation continue. On the other hand, if there is already a runnable and functional chatbot, and the only thing to do is to transfer its sentiment (or style), then the cycle gan model might be a good choice. The plug and play model might not be suitable for sentiment transfer tasks since its awkward performance.
\item In English, reinforcement Learning and CycleGAN were the most attractive.
The reinforcement learning was able to learn properly the different design goals and offer output sentences with good diversity.
The CycleGAN model primarily performed word mapping on the original response, so the output sentence quality was more or less preserved.
The Plug and Play model and Sentiment Transformation Network were not as successful at the moment, probably because it is not easy to modify the latent code of the sentences while preserving the semantics and sentence quality.
\end{enumerate}
\end{comment}

%\newpage
\bibliographystyle{IEEEbib}
\bibliography{strings,refs}

%\newpage
%\vspace{3mm}
\begin{appendices}
  \section{DATASET STATISTICS} \label{appendix:dataset}
  Below we present the details of the datasets used in different tasks. In the sentiment classifiers (Section A), there are three corpora: Chinese, PTT, and English. As described in Section~\ref{section:PTT}, the corpus used in the PTT task was refined by a special filtering processing which reduced the size from 1,402K to 1,149K. %special ???
  In the experiments (Section B), each of the three corpora are shared by four tasks~-- the persona-based model, the reinforcement learning model, the plug and play model, and the CycleGAN model. 
  In the metrics (Sections C and D), there are for metrics for the Chinese, PTT, and English tasks. The Chinese and PTT tasks share the same metrics trained by the corpora described in Section C. The WMT11 corpus used for the LM in Section D used the data pre-processing described in~\cite{chelba2013one}. 
  For word-based segmentation, low-frequency words were eliminated to reduce the word dimensions, which led to the vocabulary size shown in the table, which was smaller than the actual size.
  
  \subsection{Sentiment Classifiers}
    %\FloatBarrier
    \begin{table}[h]
    \centering
    \begin{tabular}{|P{0.75cm}|P{0.9cm}|P{0.9cm}|P{0.7cm}|P{0.75cm}|P{0.8cm}|P{0.85cm}|}
    \hline
     \multirow{3}{*}{Task} & 
     \multirow{3}{*}{Corpus} &
     \multirow{3}{*}{\begin{tabular}[c]{@{}c@{}}Training\\set\end{tabular}} &
     \multirow{3}{*}{\begin{tabular}[c]{@{}c@{}}Testing\\set\end{tabular}} &
     \multirow{3}{*}{Total} &
     \multirow{3}{*}{\begin{tabular}[c]{@{}c@{}}Vocab\\size\end{tabular}} &
     Word \\
      & & & & & & segment\\
      & & & & & & type\\
     \hhline{|=|=|=|=|=|=|=|}
     \multirow{2}{*}{Chinese} & CECG & \multirow{2}{*}{16,999K} & \multirow{2}{*}{1k} & \multirow{2}{*}{1.7M} & \multirow{2}{*}{50K} & \multirow{2}{*}{word}\\
     & \cite{CECG} & & & & & \\
     \hline
     \multirow{3}{*}{PTT} & PTT & \multirow{3}{*}{1,148K} & \multirow{3}{*}{1K} & \multirow{3}{*}{1,149K} & \multirow{3}{*}{50K} & \multirow{3}{*}{word}\\
     & \textit{Boy-Girl} & & & & & \\
     \hline
     \multirow{2}{*}{English} & Twitter & \multirow{2}{*}{14,999K} & \multirow{2}{*}{1K} & \multirow{2}{*}{15M} & \multirow{2}{*}{50K} & \multirow{2}{*}{word}\\
     & \cite{pak2010twitter} & & & & & \\
    \hline
    \end{tabular}
    \end{table}
    % not Lee but me: PTT boy-girl corpus, train, test, vocab size; English vocab size v
    % not Lee but me: PTT sentiment的corpus數駐明是經過裁減的，詳情看該單元 VI v
    
  \subsection{Experiments}
    %\FloatBarrier
    \begin{table}[h]
    \centering
    \begin{tabular}{|P{0.75cm}|P{0.9cm}|P{0.9cm}|P{0.7cm}|P{0.75cm}|P{0.8cm}|P{0.85cm}|}
    \hline
     \multirow{3}{*}{Task} & 
     \multirow{3}{*}{Corpus} &
     \multirow{3}{*}{\begin{tabular}[c]{@{}c@{}}Training\\set\end{tabular}} &
     \multirow{3}{*}{\begin{tabular}[c]{@{}c@{}}Testing\\set\end{tabular}} &
     \multirow{3}{*}{Total} &
     \multirow{3}{*}{\begin{tabular}[c]{@{}c@{}}Vocab\\size\end{tabular}} &
     Word \\
      & & & & & & segment\\
      & & & & & & type\\
     \hhline{|=|=|=|=|=|=|=|}
     \multirow{2}{*}{Chinese} & CECG & \multirow{2}{*}{16,999K} & \multirow{2}{*}{1k} & \multirow{2}{*}{1.7M} & \multirow{2}{*}{50K} & \multirow{2}{*}{word}\\
     & \cite{CECG} & & & & & \\
     \hline
     \multirow{3}{*}{PTT} & PTT & \multirow{3}{*}{1,148K} & \multirow{3}{*}{1K} & \multirow{3}{*}{1,149K} & \multirow{3}{*}{50K} & \multirow{3}{*}{word}\\
     & \textit{Boy-Girl} & & & & & \\
     \hline
     \multirow{2}{*}{English} & Twitter & \multirow{2}{*}{3,672K} & \multirow{2}{*}{28K} & \multirow{2}{*}{3.7M} & \multirow{2}{*}{50K} & \multirow{2}{*}{word}\\
     & \cite{Marsan-Ma} & & & & & \\     
    \hline
    \end{tabular}
    \end{table}
    
  \subsection{Metrics~-- Chinese \& PTT}
    %\FloatBarrier
    \begin{table}[h]
    \centering
    \begin{tabular}{|P{0.75cm}|P{0.9cm}|P{0.9cm}|P{0.7cm}|P{0.75cm}|P{0.8cm}|P{0.85cm}|}
    \hline
     \multirow{3}{*}{Task} & 
     \multirow{3}{*}{Corpus} &
     \multirow{3}{*}{\begin{tabular}[c]{@{}c@{}}Training\\set\end{tabular}} &
     \multirow{3}{*}{\begin{tabular}[c]{@{}c@{}}Testing\\set\end{tabular}} &
     \multirow{3}{*}{Total} &
     \multirow{3}{*}{\begin{tabular}[c]{@{}c@{}}Vocab\\size\end{tabular}} &
     Word \\
      & & & & & & segment\\
      & & & & & & type\\
     \hhline{|=|=|=|=|=|=|=|}
     \multirow{2}{*}{COH1} & PTT & \multirow{2}{*}{415K} & \multirow{2}{*}{1K} & \multirow{2}{*}{416K} & \multirow{2}{*}{6,185} & \multirow{2}{*}{char}\\
     & \cite{Justin-Yang} & & & & & \\
     \hline
     \multirow{2}{*}{COH2} & PTT & \multirow{2}{*}{415K} & \multirow{2}{*}{1K} & \multirow{2}{*}{416K} & \multirow{2}{*}{6,185} & \multirow{2}{*}{char}\\
     & \cite{Justin-Yang} & & & & & \\
     \hline
     \multirow{2}{*}{SCL} & CECG & \multirow{2}{*}{16,999K} & \multirow{2}{*}{1K} & \multirow{2}{*}{1.7M} & \multirow{2}{*}{50K} & \multirow{2}{*}{word}\\
     & \cite{CECG} & & & & & \\        
     \hline
     \multirow{2}{*}{LM} & PTT & \multirow{2}{*}{24,999K} & \multirow{2}{*}{1K} & \multirow{2}{*}{25M} & \multirow{2}{*}{50K} & \multirow{2}{*}{word}\\
     & \textit{Replies} & & & & & \\        
     \hline
    \end{tabular}
    \end{table}
%  \newpage  
  \subsection{Metrics~-- English}
    %\FloatBarrier
    \begin{table}[h]
    \centering
    \begin{tabular}{|P{0.7cm}|P{0.85cm}|P{1cm}|P{0.8cm}|P{0.75cm}|P{0.75cm}|P{0.85cm}|}
    \hline
     \multirow{3}{*}{Task} & 
     \multirow{3}{*}{Corpus} &
     \multirow{3}{*}{\begin{tabular}[c]{@{}c@{}}Training\\set\end{tabular}} &
     \multirow{3}{*}{\begin{tabular}[c]{@{}c@{}}Testing\\set\end{tabular}} &
     \multirow{3}{*}{Total} &
     \multirow{3}{*}{\begin{tabular}[c]{@{}c@{}}Vocab\\size\end{tabular}} &
     Word \\
      & & & & & & segment\\
      & & & & & & type\\
     \hhline{|=|=|=|=|=|=|=|}
     \multirow{2}{*}{COH1} & Twitter & \multirow{2}{*}{415K} & \multirow{2}{*}{1K} & \multirow{2}{*}{416K} & \multirow{2}{*}{6,185} & \multirow{2}{*}{char}\\
     & \cite{pak2010twitter} & & & & & \\
     \hline
     \multirow{2}{*}{COH2} & Twitter & \multirow{2}{*}{415K} & \multirow{2}{*}{1K} & \multirow{2}{*}{416K} & \multirow{2}{*}{6,185} & \multirow{2}{*}{char}\\
     & \cite{pak2010twitter} & & & & & \\
     \hline
     \multirow{2}{*}{SCL} & Twitter & \multirow{2}{*}{14,999K} & \multirow{2}{*}{1K} & \multirow{2}{*}{15M} & \multirow{2}{*}{50K} & \multirow{2}{*}{word}\\
     & \cite{pak2010twitter} & & & & & \\        
     \hline
     \multirow{2}{*}{LM} & WMT11 & 799,984K & 160K & 0.8B & \multirow{2}{*}{790K} & \multirow{2}{*}{word}\\
     & \cite{chelba2013one} & (words) & (words) & (words) & & \\        
     \hline  
    \end{tabular}
    \end{table}
  
  \section{HYPER-PARAMETER SELECTION} \label{appendix:hyper_param}
  Hyper-parameters were first chosen from the Chinese experiments and used for the subsequent PTT and English experiments. However, the epochs chosen for each task differed to ensure the best performance.
  
    \subsection{Sentiment Classifier}
      %GRU with last hidden state output(short for GRU-last)
      \begin{itemize}
          \item unit size: 256 %number of GRU unit
          \item layer size: 1
          \item batch size: 32
          \item max sequence length: 40
          \item learning rate: 0.001 (no decay)
          \item epochs: 50,000 (Chinese), 50,000 (PTT), 50,000 (English)
          \item word embedding dimension: 300
      \end{itemize}
      
    \subsection{Persona-Based Model}
      \begin{itemize}
          \item unit size of seq2seq model (both encoder and decoder): 256
          \item layer size of seq2seq model (both encoder and decoder): 1
          \item batch size: 64
          \item max sequence length: 15
          \item learning rate: 0.001 (no decay)
          \item epochs: 100,000 (Chinese), 100,000 (PTT), 100,000 (English)
          \item word embedding dimension: 300
      \end{itemize}
      
    \subsection{Reinforcement Learning Model}
      \begin{itemize}
          \item unit size of seq2seq model (both encoder and decoder): 300
          \item layer size of seq2seq model (both encoder and decoder): 4
          \item batch size: 64
          \item max sequence length: 50
          \item learning: initialized as 0.5 and decay every 500 iterations with a weight of 0.99 
          \item epochs for seq2seq model: 100,000 (Chinese), 100,000 (PTT), 100,000 (English)
          \item epochs for RL model: 2,000 (Chinese), 1,000 (PTT), 2,000 (English)
          \item word embedding dimension: 300
          \item coefficient of $R_1$, $R_2$ and $R_3$: 0.3,0.3,0.4
      \end{itemize}
      
    \subsection{Plug and Play Model}
      \begin{itemize}
          \item unit size for VAE RNN: 500*2 (bidirectional)
          \item unit size of seq2seq model (both encoder and decoder): 500
          \item layer size of seq2seq model (both encoder and decoder): 1
          \item batch size: 48
          \item max sequence length: 15
          \item learning rate: 0.001 (no decay)
          \item epochs: 40,000 (Chinese), 50,000 (PTT), 40,000 (English)
          \item word embedding dimension: 300
          \item sentiment gradient weight: 400
          \item L2 gradient weight: 25
      \end{itemize}
      
    \subsection{CycleGAN Model}
      \begin{itemize}
          \item unit size of seq2seq model (both encoder and decoder): 256
          \item layer size of seq2seq model (both encoder and decoder): 1
          \item batch size: 32
          \item max sequence length: 15
          \item learning rate: 0.0001 (no decay)
          \item epochs: 100,000 (Chinese), 100,000 (PTT), 80,000 (English)
          \item word embedding dimension: 300
          \item ratio between training iterations of discriminator and of generator: 1:1
      \end{itemize}
      
    \subsection{Metric~-- COH1}
      \begin{itemize}
          \item unit size of seq2seq model (both encoder and decoder): 300
          \item layer size of seq2seq model (both encoder and decoder): 4
          \item batch size: 32
          \item max sequence length: 50
          \item learning: initialized as 0.5 and decay every 500 iterations with a weight of 0.99 
          \item epochs: 120,000 (Chinese), 150,000 (PTT), 100,000 (English)
          \item word embedding dimension: 300
      \end{itemize}

    \subsection{Metric~-- COH2}
      \begin{itemize}
          \item unit size of seq2seq model (both encoder and decoder): [200,100,100,200]
          \item layer size of seq2seq model (both encoder and decoder): 4
          \item batch size: 64
          \item max sequence length: 30
          \item learning: initialized as 0.0005 and decay every 5000 iterations with a weight of 0.98
          \item epochs: 100,000 (Chinese), 120,000 (PTT), 100,000 (English)
          \item word embedding dimension: 300
      \end{itemize}

    \subsection{Metric~-- SCL}
      \begin{itemize}
          \item unit size of seq2seq model (both encoder and decoder): 300
          \item layer size of seq2seq model (both encoder and decoder): 3
          \item batch size: 64
          \item max sequence length: 
          \item learning: initialized as 0.0005 and decay every 5000 iterations with a weight of 0.98
          \item epochs: 50,000 (Chinese), 50,000 (PTT), 50,000 (English)
          \item word embedding dimension: 300
      \end{itemize}

    \subsection{Metric~-- LM}
      \begin{itemize}
          \item unit size: 256 %number of GRU unit
          \item layer size: 1
          \item batch size: 32
          \item max sequence length: 40
          \item learning rate: 0.001 (no decay)
          \item epochs: 50,000 (Chinese), 75,000 (PTT), 50,000 (English)
          \item word embedding dimension: 300
      \end{itemize}
      
\end{appendices}

\end{CJK*}
\end{document}